\documentclass[journal]{IEEEtran}
\usepackage{amsmath,amssymb}
\usepackage{graphicx,graphics,color,psfrag}
\usepackage{cite,balance}
 \allowdisplaybreaks
 \usepackage{algorithm}
\usepackage{algorithmic}
 \usepackage{accents}
 \usepackage{amsthm}
 \usepackage{hyperref}
 \usepackage{bm}
 \usepackage{algorithmic}
 \usepackage[english]{babel}
 \usepackage{multirow}
 \usepackage{enumerate}
 \usepackage{cases}
 \usepackage{stfloats}
 \usepackage{dsfont}
 \usepackage{color,soul}
 \usepackage{amsfonts}
 \usepackage{graphicx,amsmath,amssymb}
 \usepackage{framed}
 \usepackage{fancyhdr}
 \usepackage{hhline}
 \usepackage{ulem}
 \usepackage{graphicx,graphics}
 \usepackage{array,color}
 \usepackage{amsmath}
 \usepackage{booktabs}
 \usepackage{subfigure}
 \usepackage[nolist]{acronym}
 \usepackage{bbm}
 \usepackage{diagbox}

\makeatletter
\renewcommand{\fnum@figure}{Fig. \thefigure}
\makeatother

\newtheoremstyle{itshape}% name
  {.0\baselineskip\@plus.0\baselineskip\@minus.0\baselineskip}% Space above
  {.0\baselineskip\@plus.0\baselineskip\@minus.0\baselineskip}% Space below
  {\itshape}% Body font
  {}%Indent amount (empty = no indent, \parindent = para indent)
  {\bfseries}%  Thm head font
  {.}%       Punctuation after thm head
  { }%      Space after thm head: " " = normal interword space;
  {}%       Thm head spec

\theoremstyle{itshape}

\newtheorem{theorem}{Theorem}

\newtheorem{proposition}{Proposition}
\newtheorem{lemma}{Lemma}
\newtheorem{corollary}{Corollary}
\newtheorem{assumption}{Assumption}
\newtheorem{definition}{Definition}

\renewcommand{\algorithmicrequire}{ \textbf{Input:}}
\renewcommand{\algorithmicensure}{ \textbf{Output:}}

\begin{document}
\include{header}
\title{A Two-Timescale Approach for Wireless Federated Learning with Parameter Freezing and Power Control}

\author{Jinhao Ouyang, Yuan Liu, and Hang Liu
\thanks{Jinhao Ouyang, and Yuan Liu are with school of Electronic and Information Engineering, South China University of Technology, Guangzhou 510641, China (emails: eejhouyang@mail.scut.edu.cn, eeyliu@scut.edu.cn).\par
Hang Liu is with Department of Electrical and Computer Engineering, Cornell Tech, Cornell University, NY 10044, USA (email: hl2382@cornell.edu).\par
\emph{Corresponding author: Yuan Liu.}
}\par

}

\maketitle

\begin{abstract}
Federated learning (FL) enables distributed devices to train a shared machine learning (ML) model collaboratively while protecting their data privacy. However, the resource-limited mobile devices suffer from intensive computation-and-communication costs of model parameters. In this paper, we observe the phenomenon that the model parameters tend to be stabilized long before convergence during training process. Based on this observation, we propose a two-timescale FL framework by joint optimization of freezing stabilized parameters and controlling transmit power for the unstable parameters to balance the energy consumption and convergence. First, we analyze the impact of model parameter freezing and unreliable transmission on the convergence rate. Next, we formulate a two-timescale optimization problem of parameter freezing percentage and transmit power to minimize the model convergence error subject to the energy budget. To solve this problem, we decompose it into parallel sub-problems and decompose each sub-problem into two different timescales problems using the Lyapunov optimization method. The optimal parameter freezing and power control strategies are derived in an online fashion. Experimental results demonstrate the superiority of the proposed scheme compared with the benchmark schemes.

\end{abstract}


\section{Introduction}
The rapid proliferation of mobile devices has generated massive data, prompting the emergence of numerous machine learning (ML)-based applications, such as face recognition, augment reality, and object detection \cite{10310213}. Conventional ML approaches require centralizing the training data in a data center or cloud, leading to significant privacy concerns \cite{8755300, 8743390}. To address the issue, federated learning (FL) has emerged as a promising distributed learning paradigm that enables mobile devices to collaboratively train a shared ML model under the coordination of a central server, while protecting data privacy \cite{pmlr-v54-mcmahan17a, 9562559}. However, as ML model parameters are typically high-dimensional, the intensive local computation on the devices and the frequent communication between devices and server result in huge computation and communication overheads \cite{10304624, 9534784}, which present challenges for on-device resources as they are with limited energy budgets. \par 
There are extensive existing works aimed at tackling the issues of computation and communication efficiency in FL, and the typical methods include sparsification, quantization, and device scheduling. In sparsification, only partial elements of local gradients are uploaded, while the rest are accumulated locally \cite{10185584, 10233012, 8889996}. Quantization reduces the number of bits representing each element of model parameters, thereby reducing the total size of model parameters exchanged between devices and server \cite{NIPS2017_6c340f25, 9916128, 9272666, 10319317}. Device scheduling is used to select partial devices to participate in training so as to improve energy efficiency or convergence \cite{10285619, 9210812,10038486, 10086671}. \par
\begin{figure*}[t]
\setlength{\belowcaptionskip}{-0.4cm}
\centering
\subfigcapskip=-6pt
\subfigure[{{MNIST dataset.}}]{
    \includegraphics[clip, viewport= 5 0 420 330,width=0.3\linewidth]{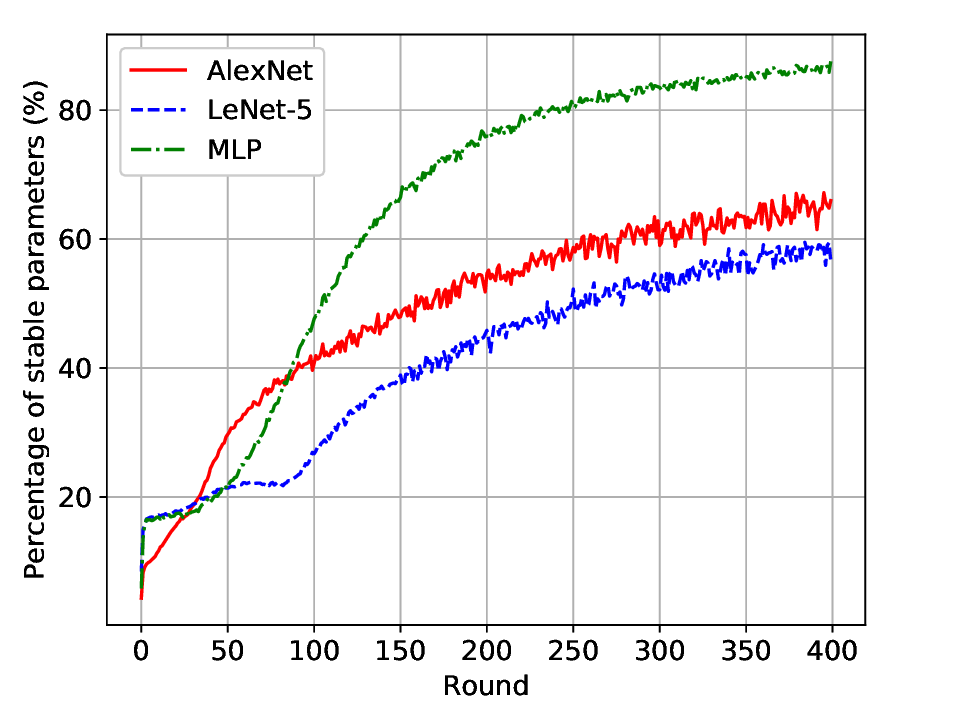}}
\ 
\subfigure[{{Fashion-MNIST dataset.}}]{
    \includegraphics[clip, viewport= 5 0 420 330,width=0.3 \linewidth]{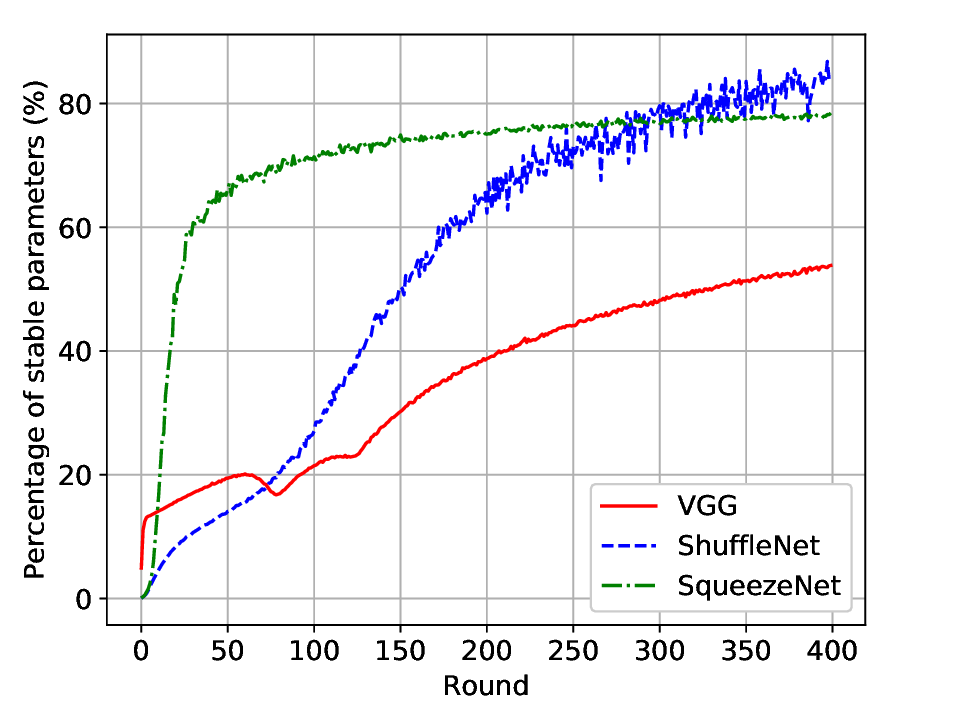}
}
\
\subfigure[{{CIFAR-10 dataset.}}]{\label{fig: Average convergence error vs E CIFAR10}
    \includegraphics[clip, viewport= 5 0 420 330,width=0.3 \linewidth]{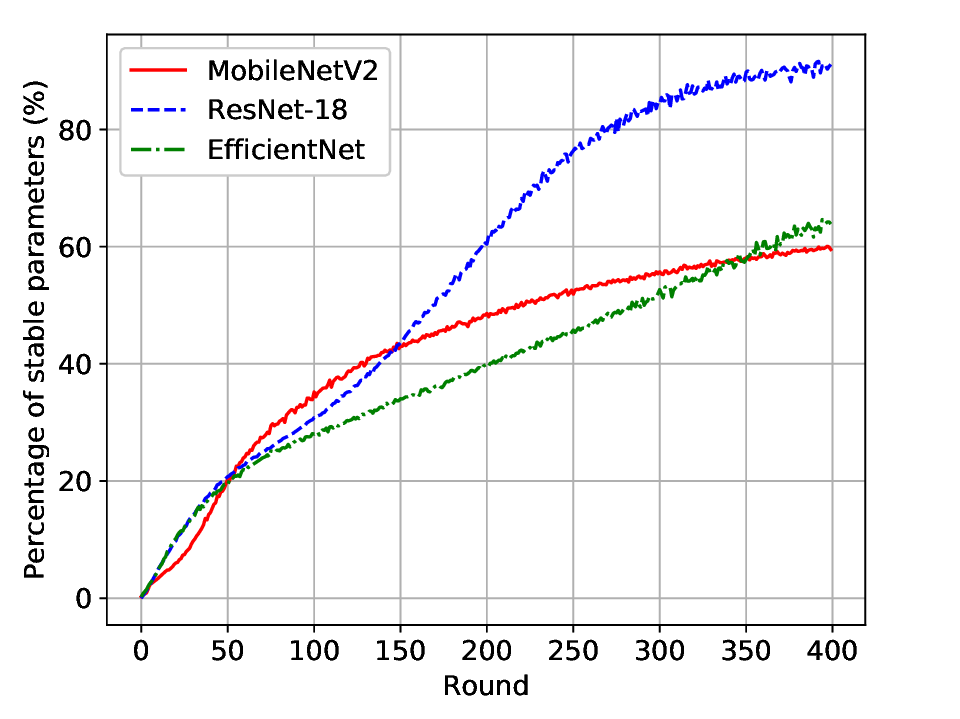}
}
\vspace{-0.12in}
\caption{{{Percentage of stable parameters versus the training round index across different image classification tasks. In these experiments, a model parameter is considered stable if its cumulative absolute update over every consecutive 
$I=10$ updates is less than 0.5\% of its total cumulative absolute update.}}} 
\label{fig: Pecentage of stable parameters}

\end{figure*}

 
Note that all the methods require each device to update the entire local model and to upload the local model with a fixed size. However, most of the model parameters tend to be stabilized long before convergence during training. To demonstrate this insight, we conduct a series of experiments on typical datasets by training various neural networks as shown in Fig. \ref{fig: Pecentage of stable parameters}, where it is observed that an increasing percentage of model parameters become stable over time. Therefore, after the parameters stabilize at their optimal values, continuously updating and uploading them becomes redundant, as it consumes extra energy in both computation and communication without improving model performance. Consequently, the stable parameters should be frozen and excluded from being updated and uploaded to save energy. Moreover, as mentioned in \cite{10036106}, the model parameters keep stable for multiple communication rounds, taking a timescale of seconds to minutes in practice (e.g. 20 $\sim$ 60 seconds in LeNet-5 and around 30 minutes in ResNet-18 training on CIFAR-10 dataset). In contrast, the unstable model parameters need to be updated and uploaded in each communication round, with the transmission time of milliseconds in general (e.g., 0.5 $\sim$ 500 milliseconds in industrial internet-of-things applications \cite{9829183}). \par
In addition to parameter freezing, various subnet training frameworks have been proposed in FL to mitigate computational and communication overhead by selectively updating a subset of model parameters. These approaches include model pruning, federated dropout, and rolling training. Adaptive model pruning, as introduced in \cite{10313248, 10678894}, reduces the neural network size during training to alleviate computational complexity. Federated dropout \cite{xie2024federated} dynamically adjusts the dropout rate to reduce the number of active parameters, thereby lowering communication and computation costs. Similarly, FjORD \cite{NEURIPS2021_6aed000a} removes adjacent components of a neural network while preserving critical parameters for training. Rolling training, as explored in \cite{NEURIPS2022_bf5311df, NEURIPS2023_52635645}, employs a rolling sub-model extraction mechanism to ensure balanced training across different parts of the model. Note that these methods may still update and upload parameters that remain stable and do not require further optimization. \par
Based on above motivation to enable energy-efficient FL, we propose to freeze the stabilized parameters in a large timescale and update/upload the unstable parameters in a small timescale. Realizing the two-timescale parameter freezing and transmission raises two challenges: (1) What are the appropriate percentages of model parameters to be frozen during the training process? As shown in Fig. \ref{fig: Pecentage of stable parameters}, the percentages of stable parameters increase with training rounds, thus the freezing percentages should be dynamic over the training process; Moreover, the freezing percentages directly determine the computation and communication loads. (2) What is the optimal transmission power policy of devices? Due to the fact that the sizes of model parameters are different in large-timescale and the channel conditions vary in small-timescale, the devices need perform real-time power control to transmit model parameters over training process using limited energy budget. The need to address these two issues motivates this paper. \par
In this paper, we propose a two-timescale FL framework for joint parameter freezing percentage optimization and power control to strike a balance between energy consumption and learning performance over wireless networks. The main contributions of this paper are summarized as follows: 
\begin{itemize}
    \item We formulate an online two-timescale optimization problem with joint parameter freezing (in large-timescale) and transmit power control (in small-timescale). Our goal is to minimize the model convergence error subject to the energy budgets of mobile devices. 
    \item By using the Lyapunov optimization method, the original problem is decomposed into parallel sub-problems, with each being decoupled into two different timescale problems. We then derive the optimal parameter freezing percentages and power control strategies. 
    \item Several useful insights are obtained via our analysis: First, freezing more parameters reduces the transmission bits but allows more devices to participate in training, which can accelerate model convergence. However, if the parameter freezing percentage exceeds a certain threshold, model performance degrades due to an increase error. Second, a larger energy budget allows for a smaller parameter freezing percentage or tolerates higher transmit power. Third, the optimal power transmission policies of devices follow a threshold structure, i.e., in each slot, a device either transmits with a minimum power level or drops out the training. 
\end{itemize}\par
The rest of the paper is organized as follows. We introduce the system model in Section \ref{sec:system_model}. We analyze the convergence rate of the proposed FL framework and formulate the optimization problem in Section \ref{sec:convergence_analysis and problem formulation}. The problem solution is presented in Section \ref{sec:Problem Solution} and the experimental results are provided in Section \ref{sec:Experimental_Results}. In Section \ref{sec:Conclusion}, we conclude the paper.
\section {System Model}
\label{sec:system_model}
In this section, we provide an overview of the considered system model and introduce the technical preliminaries used in this paper. 
\begin{figure}[t]
\centering
\includegraphics[width=0.45\textwidth]{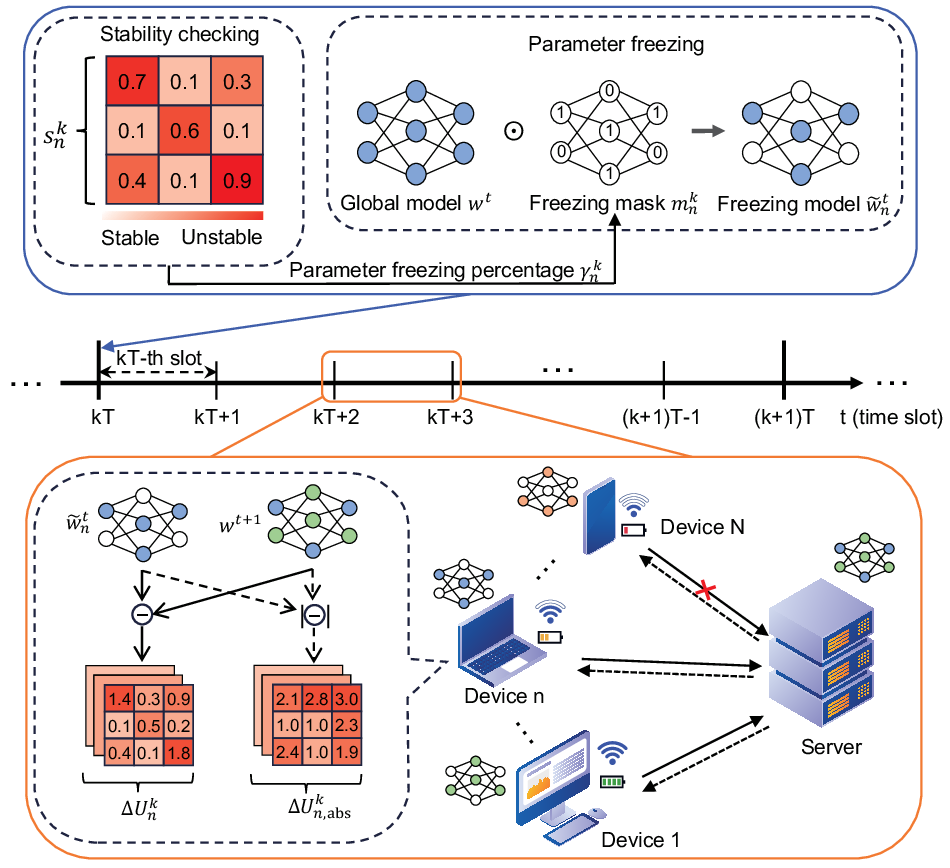}
\caption{Two-timescale FL system.}
\label{fig FL system}
\vspace{-0.5cm}
\end{figure}
\subsection{Two-Timescale FL Structure}
As shown in Fig. \ref{fig FL system}, we consider a wireless FL system comprising one server and a set of distributed devices, denoted by $\mathcal N = \{1, 2, \cdots, N\}$. Note that in FL, local gradient uploading is performed in each communication round, while parameter freezing tends to last for multiple communication rounds. In this regard, we refer to a single communication round as a slot, indexed by $t \in \{0, 1, \cdots \}$, group every consecutive $T$ time slots as a frame, indexed by $k \in \{0, 1, \cdots \}$, and denote the set of time slots in the $k$-th frame as $\mathcal{T}_k = \{kT, kT + 1, \cdots, (k + 1)T-1\}$. \par
At time slot $t = kT$, i.e., the beginning of a frame, each device freezes the stable parameters from both training and uploading for the $k$-th frame, i.e., 
\begin{align}\label{freezing operation}
    \tilde{\bm{w}}_{n}^{t} = \bm{w}^{t} \odot \bm{m}_{n}^{k}, 
\end{align}
where $\tilde{\bm{w}}_{n}^{t}$ and $\bm{w}^{t}$ represent the local freezing model of device $n$ and the global model broadcast by server, respectively, $\bm{m}_{n}^{k}$ is the mask vector of parameter freezing for device $n$ in the $k$-th frame and $\odot$ represents the freezing operation. Note that the operator $\odot$ in (\ref{freezing operation}) denotes a masking operation, which uses $\bm{m}_{n}^{k}$ to indicate the parameters to be frozen, rather than representing the Hadamard product. \par
Then in slot $t \in \mathcal{T}_k$, each device trains the local freezing model based on its local data, and the local gradient is computed as 
\begin{align} \label{eqn:local gradient train}
    \nabla F_n(\tilde{\bm{w}}_{n}^{t}) = \frac{1}{B_n^t} \sum_{i=1}^{B_n^t} \nabla F_n(\tilde{\bm{w}}_{n}^{t}; \bm{\zeta}_{n}^{i}, \bm{w}_n^t), 
\end{align}
where $B_{n}^t$ is the total training data size of device $n$ at slot $t$, $\bm{w}_{n}^{t}$ represents the local model without parameter freezing, and $\nabla F_n(\tilde{\bm{w}}_{n}^{t}; \bm{\zeta}_{n}^{i}, \bm{w}_n^t)$ is the local gradient computed from data sample $\bm{\zeta}_{n}^{i}$ with model parameters $\tilde{\bm{w}}_{n}^{t}$. For simplicity, we assume that the training data size remains the same for each slot throughout the learning process, i.e., $B_{n}^t = B_{n}, \forall n \in \mathcal{N}, t = 0, 1, \cdots$. \par

After local training, each device uploads the local gradient parameters to the server. Due to the time-varying channel conditions, transmission outage may occur when the channel exhibits deep fading and the transmission requirement cannot be met. In this paper, we define transmission outage as follows. 
\begin{definition}\label{definition outage}
(Transmission outage): For any device $n$, the transmission outage occurs when the sum of communication latency $\tau_{n}^{\rm{com}}$ and computation latency $\tau_{n}^{\rm{cmp}}$ exceeds a  given per-round latency $\tau_0$, i.e, $\tau_{n}^{\rm{cmp}}+\tau_{n}^{\rm{com}} > \tau_0$. 
\end{definition}\par 
Thus, after every $\tau_0$ interval, the server aggregates the local gradients as follows: 
\begin{align}\label{eqn:global gradient aggregation}
     g(\tilde{\bm{w}}^{t}) = \frac{\sum_{n=1}^{N} \mathbbm {1}_{n}^{t} B_{n}^t  \nabla{F_{n}(\tilde{\bm{w}}_{n}^{t})}}{\sum_{n=1}^{N} \mathbbm {1}_{n}^{t} B_{n}^t }, 
\end{align}
where $g(\tilde{\bm{w}}^{t})$ is the global gradient and $\mathbbm {1}_{n}^{t}\in \{0,1\}$ is an indicator function. $\mathbbm {1}_{n}^{t}= 0$ indicates that the transmission outage occurred and the server cannot  correctly receive the local gradient of device $n$. Specifically, the indicator function can be defined as 
\begin{align}\label{indicator function}
\mathbbm {1}_{n}^{t} = 
\begin{cases} 
1,  &\ \text{if } \tau_{n}^{{\rm cmp},t}+\tau_{n}^{{\rm com},t} \leq \tau_0; \\
0,  &\ \text{otherwise}.
\end {cases}
\end{align}\par
Then the server updates the global model by 
\begin{align}\label{eqn:global model update}
    \bm{w}^{t+1} = \bm{w}^{t} - \eta g(\tilde{\bm{w}}^{t}), 
\end{align}
where $\eta >0$ is the learning rate. \par
At the end of each slot, the server broadcasts the updated global model $\bm{w}^{t+1}$ to all devices, and each device $n$ calculates the discrepancy between the local freezing model $\tilde{\bm{w}}_{n}^{t}$ and the updated global model $\bm{w}^{t+1}$ to accumulate the residual as a metric for checking parameter stability. Specifically, let $\Delta \bm{U}_{n}^{k, t}$ and $\Delta \bm{U}_{n, \rm{abs}}^{k, t}$ denote the cumulative update and the cumulative absolute value of the update of device $n$ at slot $t \in \mathcal{T}_k$ in $k$-th frame, we have 
\begin{align}\label{eqn:cumulative update}
\begin{cases}
    \Delta \bm{U}_{n}^{k, t} = \Delta \bm{U}_{n}^{k, t-1} + \tilde{\bm{w}}_{n}^{t} - \bm{w}^{t+1}, \\
    \Delta \bm{U}_{n, \rm{abs}}^{k, t} = \Delta \bm{U}_{n, \rm{abs}}^{k, t-1} + |\tilde{\bm{w}}_{n}^{t} - \bm{w}^{t+1}|,
\end{cases}
\end{align}
where $|\bm{x}|$ denotes the absolute value taken for each element in the vector $\bm{x}$. And at the beginning of the $k$-th frame, we let $\Delta \bm{U}_{n}^{k, t} =  \Delta \bm{U}_{n, \rm{abs}}^{k, t} = 0$. 
Then at the end of  the $k$-th frame, the parameter stability vector of device $n$, denoted by a $D$-dimensional vector $\bm{s}_n^k$, can be calculated as 
\begin{align}
    \bm{s}_n^k = \frac{|\Delta \bm{U}_{n}^{k, t}|}{\Delta \bm{U}_{n, \rm{abs}}^{k, t}}, {\text{if}}\  t = (k+1)T-1.  
\end{align}
Note that each element $s_n^k[d]$ in $\bm{s}_n^k$ individually represents the stability of a parameter with $s_n^k[d] \in [0, 1]$, where $d$ is an index in $\{1, 2, \cdots, D\}$. As mentioned in \cite{10036106}, a stable parameter oscillates slightly around its stationary point, suggesting that two consecutive model updates well counteract each other. In contrast, an unstable parameter means that the model updates move in the same direction towards its stationary point across multiple slots within a single frame. Therefore, as $s_n^k[d]$ approaches to 0, the corresponding parameter becomes increasingly stable, and vice versa. \par 
At the beginning of the next frame (i.e., the $(k+1)$-th frame), each device sorts $\bm{s}_n^k$ obtained in the previous frame and then computes $\bm{m}_n^{k+1}$ by determining the masking operation threshold, which is constrained by the parameter freezing percentage of the $(k+1)$-th frame $\gamma_n^{k+1}$. Finally, each device freezes the parameters for the $(k+1)$-th frame according to $\bm{m}_n^{k+1}$. The optimization of the designing variable $\gamma_n^{k+1}$ will be discussed in Section \ref{sec:Problem Solution}.\par
In summary, the processes of the proposed two-timescale FL framework within a frame can be described as follows: 
\begin{itemize}
    \item{\bf Per frame (large timescale) operation:} Parameter stability checking and parameter freezing are conducted at the beginning of each frame and the parameter freezing percentage remains unchanged during a frame. 
    \item{\bf Per slot (small timescale) operation:} Each device decides whether to train the local freezing model as (\ref{eqn:local gradient train}) and update the local gradient to the server based on the transmit power control strategy performed at each time slot. Subsequently, the server aggregates the received local gradients as (\ref{eqn:global gradient aggregation}), updates the global model as (\ref{eqn:global model update}) and broadcasts the updated model to all devices. Finally, each device calculates the cumulative update of the local model as (\ref{eqn:cumulative update}) for parameter stability checking. 
\end{itemize}

\subsection{Communication Model}
The achievable rate of device $n$ at slot $t$ is given by
\begin{align}\label{eqn:com_rate}
\textit{r}_{n}^{t}= W\log_2(1+\frac{p_{n}^{t} h_{n}^{t}}{{N}_0}), 
\end{align}
where $p_{n}^{t}$ is the transmit power of device $n$, $W$ is the channel bandwidth, $h_{n}^{t}$ is the uplink channel power gain between device $n$ and server, and $\textit{N}_0$ represents the noise power. Then given a parameter freezing percentage $\gamma_n^k$, the corresponding communication latency and energy consumption at the $t$-th slot in the $k$-th frame are given by
\begin{align}\label{eqn:com_delay}
\tau_{n}^{{\rm com},t} = \frac{(1-\gamma_n^k)S}{r_n^t}, 
\end{align}
\begin{align}\label{eqn:com_energy}
{E}_{n}^{{\rm com},t} = \frac{p_{n}^{t} (1-\gamma_n^k)S}{r_n^t}, 
\end{align}
in which $S$ is the size of the entire gradient parameters without freezing (in bits). 

\subsection{Computation Model} 
Denote $f_n$ (in cycles per second) as the computation frequency of device $n$, and $c_n$ as the number of CPU cycles required by device $n$ to process each sample. For simplicity, we assume that $c_n$ and $f_n$ are constant throughout the training process \cite{9673130}. Then, given the parameter freezing percentage $\gamma_n^k$, the corresponding computation latency and energy consumption at the $t$-th slot in the $k$-th frame are given by
\begin{align}\label{eqn:cmp_time}
\tau_{n}^{{\rm cmp},t} = \frac{(1-\gamma_n^k)c_n B_{n}^t}{f_n}, 
\end{align}
\begin{align}\label{eqn:cmp_energy}
{E}_{n}^{{\rm cmp},t} = \frac{\alpha_{n}}{2} (1-\gamma_n^k)c_n B_{n}^t f_{n}^{2}, 
\end{align}
where $\alpha_n$ is the effective capacitance coefficient depending on the chip of device $n$. \par
Therefore, the total energy consumption of a device is the sum of its communication and computation energy, which is given by
\begin{align} \label{client energy consumtion}
    E_n^{t} = E_{n}^{{\rm cmp},t}  + E_{n}^{{\rm com},t}, \forall n \in \mathcal{N}. 
\end{align}
\section{Convergence Analysis And Problem Formulation }
In this section, we first conduct a convergence analysis of the proposed FL scheme. Then a joint optimization problem of parameter freezing percentage and transmit power control is formulated to minimize the model convergence error subject to the energy budget.
\label{sec:convergence_analysis and problem formulation}
\subsection{Convergence Analysis}
\label{sec:convergence_analysis}
In this subsection, through convergence analysis, we aim to reveal how the parameter freezing and transmission outage jointly affect the convergence error. We consider a smooth non-convex learning problem with the following assumptions.
\begin{assumption}\label{Assumption 1}
\textit{(L-smoothness)}: The gradient of each local loss function $F_n$ is Lipschitz continuous with a positive constant $L$ for each of device $n\in \mathcal{N}$, i.e., $\forall \bm{v}, \bm{w} \in \mathbb{R}^d$, $\Vert\nabla{F}_n(\bm{v})-\nabla{F}_n(\bm{w})\Vert  \leq L\Vert\bm{v}-\bm{w}\Vert$.
\end{assumption}
\begin{assumption}\label{Assumption 3}
\textit{(Bounded gradient)}: For any device $n$, $\exists \xi_1, \xi_2 \geq 0$, so that the squared norm of gradient is bounded, i.e., $ \Vert\nabla{F}_n(\bm{w})\Vert^2 \leq \xi_1 +\xi_2 \Vert\nabla{F}(\bm{w})\Vert^2$.
\end{assumption}
\begin{assumption}\label{Assumption 4}
\textit{(Bounded parameter gap induced by freezing)}: The norm of the parameter gap induced by parameter freezing is uniformly upper bounded by $\sigma^2$ throughout the learning process, i.e., $\Vert\bm{w}_n^t - \tilde{\bm{w}}_{n, \text{full}}^t \Vert^2 \leq \sigma^2, \forall n, t$, where $\tilde{\bm{w}}_{n, \text{full}}^t$ denotes the model with full parameter freezing, and $\bm{w}_n^t$ denotes the model updated without parameter freezing. 
\end{assumption}
\begin{assumption}\label{Assumption 5}
\textit{(Bounded data variance)}: For any device $n$, $\mathbb{E}\left[\Vert \nabla F_n(\bm{w}) - \nabla F(\bm{w}) \Vert^2\right] \leq \mathcal{X}_n^2$, which measures the heterogeneity of local datasets.
\end{assumption}\par
The above assumptions are widely used in the literature of convergence analysis for FL \cite{10443546, gu2021fast, 10535531, 9210812, 10313248, 10050151, 9611373}. Moreover, due to the fact that frozen parameters are not updated, there exists a gap between models updated with and without parameter freezing. Thus, similar to \cite{10036106}, we employ Assumption \ref{Assumption 4} to bound this gap. Here, we highlight that even when all model parameters are frozen, the gap introduced by freezing remains bounded. Assumption \ref{Assumption 4} has been theoretically justified in \cite{10036106}. To further validate this assumption, two supporting experiments are provided in Section \ref{Justification of Assumption 3}. Then we have the following lemma.
\begin{lemma}\label{lemma 1}
For any model $\bm{w}_{n}^{t}, \tilde{\bm{w}}_{n}^{t} \in \mathbb{R}^D$, given a parameter freezing percentage $0 \leq \gamma \leq 1 $, it holds that, 
\begin{align}
    \mathbb{E}\left[ \Vert\bm{w}_{n}^{t} - \tilde{\bm{w}}_{n}^{t}\Vert^2\right] \leq \gamma \sigma^2. 
\end{align} 
where $\tilde{\bm{w}}_{n}^{t}$ represents the model updated with parameter freezing, $\bm{w}_{n}^{t}$ represents the local model without parameter freezing. Here, the expectation is taken over the randomness in the selection of frozen parameters.
\end{lemma}
\begin{proof}
    According to \cite{10036106}, for those non-frozen parameters that are updated regularly, there is no gap incurred. Only the frozen parameters, which are not updated, incur a model parameter gap. Moreover, in the proposed FL framework, more stable parameters are prioritized for freezing based on the sorted parameter stability vector $\bm{s}_n^k$. Accordingly, $\gamma$ percent of the parameters with the smallest parameter gap are selected for freezing. The following definition is then introduced. 
    \begin{definition}
    For a parameter $1 \leq d \leq D$, the parameter freezing operators are defined for $\bm{w}_n^t \in \mathbb{R}^D$ as
\begin{align}
    (\bm{w}_n^t - \tilde{\bm{w}}_n^t)_{d} :=
\begin{cases}
    (\bm{w}_n^t - \tilde{\bm{w}}_n^t)_{\pi_{(d)}}, & \text{if } d \leq \lfloor \gamma \cdot D \rfloor, \\
    0, & \text{otherwise},
\end{cases}
\end{align}
\begin{align}
(\bm{w}_{n}^{t} - {\rm{rand}}_{\gamma}(\tilde{\bm{w}}_{n}^{t}))_d :=
\begin{cases}
    (\bm{w}_n^t - \tilde{\bm{w}}_n^t)_d, & \text{if } d \in \omega_{\gamma}, \\
    0, & \text{otherwise},
\end{cases}
\end{align}
where $(\bm{x})_d$ denotes the $d$-th element of vector $\bm{x}$, $\lfloor \cdot \rfloor$ denotes the floor operation, $\tilde{\bm{w}}_n^t$ denotes the local model obtained by freezing the $\gamma$ percent of parameters with the smallest gap, and ${\rm{rand}}_{\gamma}(\tilde{\bm{w}}_{n}^{t})$ denotes the local model obtained by randomly freezing $\gamma$ percent of parameters. 
$\pi$ is a permutation of $(\bm{w}_n^t - \tilde{\bm{w}}_n^t)$ such that, $(\left|\bm{w}_n^t - \tilde{\bm{w}}_n^t\right|)_{\pi_{(d)}} \leq (\left|\bm{w}_n^t - \tilde{\bm{w}}_n^t\right|)_{\pi_{(d+1)}}$ for $d = 1, 2, \cdots, D-1$. The set $\Omega_{\gamma}$ consists of all $\gamma$-percent subsets of model parameters selected uniformly at random. Additionally, $\omega_{\gamma}$ is a randomly selected subset from $\Omega_{\gamma}$, i.e., $\omega_{\gamma} \in \Omega_{\gamma}$ and $\omega_{\gamma} \sim_{{\rm{u.a.r}}} \Omega_{\gamma}$. 
\end{definition}
Based on the above definition, it follows that  
\begin{align}
    \Vert \bm{w}_n^t - \tilde{\bm{w}}_n^t \Vert^2 \leq \Vert \bm{w}_{n}^{t} - {\rm{rand}}_{\gamma}(\tilde{\bm{w}}_{n}^{t}) \Vert^2. 
\end{align}
This inequality arises from the fact that $\gamma$ percent of the parameters with the smallest gap are selected for freezing. Then taking the expectation over the randomness in the selection of frozen parameters, we obtain 
\begin{align}
    &\mathbb{E} \left[ \Vert \bm{w}_n^t - \tilde{\bm{w}}_n^t \Vert^2\right] \nonumber\\
    & \leq \mathbb{E} \left[\Vert \bm{w}_{n}^{t} - {\rm{rand}}_{\gamma}(\tilde{\bm{w}}_{n}^{t}) \Vert^2 \right]\nonumber\\
    &= \mathbb{E}_{\omega_{\gamma}} \left[ \Vert \bm{w}_{n}^{t} - {\rm{rand}}_{\gamma}(\tilde{\bm{w}}_{n}^{t}) \Vert^2 \right] \nonumber\\
    &= \frac{1}{|\Omega_{\gamma}|} \sum_{\omega_{\gamma} \in \Omega_{\gamma}} \sum_{d=1}^{D} \left(\bm{w}_n^t - \tilde{\bm{w}}_{n, {\rm{full}}}^t\right)_{d}^2 \mathbb{E}_{\omega_{\gamma}} [\mathbbm{1}_{\{d \in \omega_{\gamma}\}}]\nonumber\\
    & = \sum_{d=1}^{D} \left(\bm{w}_n^t - \tilde{\bm{w}}_{n, {\rm{full}}}^t\right)_{d}^2 \sum_{\omega_{\gamma} \in \Omega_{\gamma}} \frac{\mathbb{E}_{\omega_{\gamma}} [\mathbbm{1}_{\{d \in \omega_{\gamma}\}}]}{|\Omega_{\gamma}|}\nonumber\\
    & = \gamma \sum_{d=1}^{D} \left(\bm{w}_n^t - \tilde{\bm{w}}_{n, {\rm{full}}}^t\right)_{d}^2 \nonumber\\
    &\overset{(a)}{\leq} \gamma \sigma^2, 
\end{align}
where $\mathbb{E}_{\omega_{\gamma}} [\mathbbm{1}_{\{d \in \omega_{\gamma}\}}] = \gamma$ and inequality (a) follows the Assumption \ref{Assumption 4}. We complete the proof. 
\end{proof} 
The main convergence result is stated as below.
\begin{theorem}\label{theorem1}
    For the considered FL scheme, the expected convergence error in communication round $t$ of the $k$-th frame, defined as $\mathbb{E}[\Vert \nabla F(\bm{w}^{t})\Vert^2]$, can be bounded as follows:
\begin{align}\label{the expected convergence error}
    &\mathbb{E}[\Vert\nabla F(\bm{w}^{t})\Vert^2] \nonumber\\ 
    &\leq \frac{2L}{1-9\xi_2} \mathbb{E}\left[F(\bm{w}^t)-F(\bm{w}^{t+1})\right] + \underbrace{\frac{3}{(1-9\xi_2)B^t}\sum\limits_{n=1}^{N} B_n^t \mathcal{X}_n^2}_{\textit{error caused by data heterogeneity}}\nonumber \\
    &+ \underbrace{\frac{6L^2 \sigma^2}{(1-9\xi_2)B^t}\sum\limits_{n=1}^{N} \mathbbm{1}_n^t B_{n}^t \gamma_n^k}_{X_1^t \textit{ caused by parameter freezing}} +\underbrace{\frac{9\xi_1}{(1-9\xi_2)B^t}\sum\limits_{n=1}^{N} (1-\mathbbm{1}_n^t) B_{n}^t}_{X_2^t \textit{ caused by transmission outage}}, 
\end{align}
where $B^t = \sum_{n=1}^{N} B_{n}^t$ is the total data size of all devices.
\end{theorem}
\begin{proof}
    See Appendix A. 
\end{proof}
From Theorem \ref{theorem1}, we observe that a reduction in the parameter freezing percentage $\gamma_n^k$ results in a diminishing model convergence error. However, this can also increase the consumption of communication and computation resources as more parameters are involved in both updating and transmission. On the other hand, increasing transmit power reduces transmission outage and enables more devices to participate successfully in training (i.e., enforcing $\mathbbm{1}_n^t$ to approach one), but this also leads to higher energy consumption. Therefore, it is crucial to determine an appropriate parameter freezing percentage and transmit power to balance the learning performance and energy consumption during the FL process. 
\begin{corollary}\label{corollary 1}
Let Assumptions \ref{Assumption 1}-\ref{Assumption 5} hold, the $KT$-rounds convergence error bound is given by 
\begin{align}\label{long-term-upper-bound}
    &\frac{1}{KT} \sum_{k=0}^{K-1}\sum_{t \in \mathcal{T}_k}\mathbb{E}[\Vert\nabla F(\bm{w}^{t})\Vert^2] \nonumber\\ 
    &\leq \frac{2L \mathbb{E}\left[F(\bm{w}^0)-F(\bm{w}^{*})\right]}{KT(1-9\xi_2)} \nonumber\\ &+ \frac{3}{KT(1-9\xi_2)B^t} \sum_{k=0}^{K-1}\sum_{t \in \mathcal{T}_k}\sum\limits_{n=1}^{N} B_n^t \mathcal{X}_n^2\nonumber \\
    &+ \frac{6L^2 \sigma^2}{KT(1-9\xi_2)B^t}\sum_{k=0}^{K-1}\sum_{t \in \mathcal{T}_k}\sum\limits_{n=1}^{N} \mathbbm{1}_n^t B_{n}^t \gamma_n^k \nonumber\\
    &+\frac{9\xi_1}{KT(1-9\xi_2)B^t}\sum_{k=0}^{K-1}\sum_{t \in \mathcal{T}_k}\sum\limits_{n=1}^{N} (1-\mathbbm{1}_n^t) B_{n}^t,
\end{align}
where $\bm{w}^*$ denotes the optimal global model parameter that minimizes the global objective function $F(\bm{w})$ in the proposed FL framework, i.e., $\bm{w}^* = \arg \min\limits_{\bm{w}} F(\bm{w})$. 
\end{corollary}
From Corollary \ref{corollary 1}, we observe that the latter two terms on the right-hand side (R.H.S.) of (\ref{long-term-upper-bound}) are highly related to parameter freezing and transmission outage. To minimize the model convergence error, one can minimize these two terms by jointly designing the parameter freezing and transmission strategies. However, directly minimizing these terms is impractical because it requires obtaining the devices’ channel state information throughout the entire training process at the start of FL. Therefore, based on Theorem \ref{theorem1}, we decouple the long-term problem into a two-timescale training round level and apply a Lyapunov-based method to enhance long-term performance.
\begin{corollary}\label{corollary 2}
Under the assumption that the CSI is unknown, suppose that the channel power gain $h_n^t \sim \exp\left(\frac{1}{H_n^t}\right)$ is independent and identically distributed (i.i.d.) over the slots of each device. Then, we have 
\begin{align}\label{the expected convergence error 2}
    &\mathbb{E}[\Vert\nabla F(\bm{w}^{t})\Vert^2] \nonumber\\
    &{\leq} \frac{2L}{1-9\xi_2} \mathbb{E}\left[F(\bm{w}^t)-F(\bm{w}^{t+1})\right] + \frac{3}{(1-9\xi_2)B^t}\sum\limits_{n=1}^{N} B_n^t \mathcal{X}_n^2 \nonumber \\
    &+ \frac{6L^2 \sigma^2}{(1-9\xi_2)B^t}\sum\limits_{n=1}^{N} (1- q_n^t) B_{n}^t \gamma_n^k +\frac{9\xi_1}{(1-9\xi_2)B^t}\sum\limits_{n=1}^{N} q_n^t B_{n}^t, 
\end{align}
where $q_n^t = 1 - e^{-\frac{h_{n, \min}^t}{H_n^t}}$ is the transmission outage probability, $h_{n, \min}^t = \frac{N_0}{p_n^t}\left( 2^{\frac{(1-\gamma_n^k)S}{W(\tau_0 - \tau_n^{{\rm{cmp}}, t})}} - 1\right)$, $\tau_{n}^{{\rm cmp},t} = \frac{(1-\gamma_n^k)c_n B_{n}^t}{f_n}$, $H_n^t$ is the large-scale fading.
\end{corollary}
From Corollary \ref{corollary 2}, we observe that the transmission outage probability is jointly influenced by the parameter freezing percentage and the transmit power.  On one hand, a reduction in the parameter freezing percentage $\gamma_n^k$ increases the transmission outage probability. This is because, as indicated by Equations (\ref{eqn:com_delay}) and (\ref{eqn:cmp_time}), a reduction in $\gamma_n^k$ increases the computation and communication latency required to update and upload a larger portion of model parameters, making it less likely to complete the training within the given per-round latency $\tau_0$. This in turn increases the model convergence error due to transmission outage. At the same time, as indicated by the third term on the R.H.S. of (\ref{the expected convergence error 2}), a reduction in the parameter freezing percentage $\gamma_n^k$ decreases the model convergence error associated with parameter freezing. However, this improvement in model convergence comes at the cost of increased energy consumption, as more resources are required to update and upload the unfrozen model parameters. On the other hand, increasing the transmit power reduces the transmission outage probability but also results in higher energy consumption. Thus, a trade-off exists between minimizing model convergence error and energy consumption, which can be managed by jointly optimizing the parameter freezing percentage and transmit power.   
\begin{proof}
    See Appendix B. 
\end{proof}

\subsection{Problem Formulation}
\label{sec:problem formulation}
Based on the proposed two-timescale FL framework, we aim to design an online parameter freezing and power control algorithm that minimizes the convergence error while continuously satisfying the constraint of the limited energy budget. Specifically, the algorithm monitors the convergence error and energy consumption in real-time, dynamically adjusting the parameter freezing percentage and transmit power to achieve the trade-off between improving model performance and reducing energy consumption. \par
Note that obtaining the exact expression for the expected convergence error is intractable. Therefore, we derive an upper bound in Theorem \ref{theorem1}, which provides theoretical guidance for parameter freezing and power control strategies \cite{9598845}. By ignoring constant terms, minimizing the upper bound of the expected convergence error in communication round $t$ is equivalent to minimizing the following term: 
\begin{align}
    X^t &= \sum\limits_{n=1}^{N} X_n^t = \lambda \sum\limits_{n=1}^{N} \mathbbm{1}_n^t B_{n}^t (\gamma_n^k-1), 
\end{align}
where $\lambda =  \max\{\frac{6L^2 \sigma^2}{(1-9\xi_2)B^t}, \frac{9\xi_1}{(1-9\xi_2)B^t}\}$. \par 
In wireless FL, devices typically have finite energy budgets due to the limited battery capacities. Consequently, they must carefully manage energy consumption in each communication round so that they can participate in the entire training process as much as possible. In this regard, we impose the following constraint on the energy consumption for each device: 
\begin{align} \label{energy constraint}
    \lim_{K \to \infty} \frac{1}{KT} \sum_{k=0}^{K-1} \sum_{t \in \mathcal{T}_k} \mathbb{E} \left \{E_n^t\right \} \leq \overline{E}_n, \forall n \in \mathcal{N}, 
\end{align}
where $\overline{E}_n$ is the pre-determined energy consumption threshold of device $n$, which can be seen as the reliability requirement of energy consumption. The expectation $\mathbb{E} \left \{ \cdot \right \}$ is taken over the randomness of channel conditions. \par
Our goal is to optimize both parameter freezing percentage and transmit power of each device to minimize the long-term model convergence error of the proposed FL scheme, subject to the average energy budget of (\ref{energy constraint}). Thus the optimization problem is formulated as \\
$\noindent\mathcal{P}1:$
\begin{subequations}\label{problem 1}
    \begin{align}
        \min_{\{p_{n}^{t}, \gamma_{n}^{k}\}} \quad & X_{\text{av}} \triangleq \lim_{{K \to \infty}} \frac{1}{KT} \sum_{k=0}^{K-1} \sum_{t \in \mathcal{T}_k} \mathbb{E} \left\{X^t\right\} \\
        \text{s.t.} \quad &  E_{\text{av}} \triangleq \lim_{{K \to \infty}} \frac{1}{KT} \sum_{k=0}^{K-1} \sum_{t \in \mathcal{T}_k} \mathbb{E} \left\{E_n^t\right\} \leq \overline{E}_n, \forall n \in \mathcal{N}, \label{constraint 1} \\
        & 0 \leq \gamma_{n}^{k} \leq 1, \forall n \in \mathcal{N}, k = 0,1,\cdots  \label{constraint 2}\\ 
        & 0 \leq p_{n}^{t} \leq \overline{P}_n, \forall n \in \mathcal{N}, t = 0,1,\cdots  \label{constraint 3}
    \end{align}
\end{subequations}
where $\overline{P}_n$ is the peak power constraint for device $n$. \par
There are two major challenges in solving Problem $\mathcal{P}1$. First, achieving optimal solutions for Problem $\mathcal{P}1$ requires complete channel state information for all devices throughout the entire training process, which is impractical to acquire in advance. Second, the parameter freezing percentage {$\gamma_{n}^{k}$} and the power allocation {$p_{n}^{t}$}, which vary across different timescales, are tightly coupled. For instance, the parameter freezing percentage {$\gamma_{n}^{k}$} for the $k$-th frame impacts the power allocations {$p_{n}^{t}$} in slots $t \in \mathcal{T}_k$, and vice versa. To address these challenges, we develop an online two-timescale control algorithm in the following section. 

\section{Problem Solution}\label{sec:Problem Solution}
In this section, we present the framework design of the proposed online algorithm. First, we decompose Problem $\mathcal{P}1$ into $N$ parallel sub-problems. Second, we transform each sub-problem into an online optimization problem using the Lyapunov optimization technique. Third, a two-timescale control algorithm is designed to solve the transformed problems optimally.
\vspace{-0.1cm}
\subsection{Problem Decomposition and Transformation} 
% \vspace{-0.1cm}
Since in the proposed FL framework, each device makes parameter freezing and power control decisions independently, Problem $\mathcal{P}1$ can be decomposed into $N$ parallel sub-problems. Specifically, for any device $n \in \mathcal{N}$, we have \\
$\noindent\mathcal{P}2:$
\begin{subequations}\label{problem 2}
    \begin{align}
        \min_{\{p_{n}^{t}, \gamma_{n}^{k}\}} \quad & X_{n,\text{av}} \triangleq \lim_{{K \to \infty}} \frac{1}{KT} \sum_{k=0}^{K-1} \sum_{t \in \mathcal{T}_k} \mathbb{E} \left\{X_n^t\right\} \\
        \text{s.t.} \quad &  E_{n,\text{av}} \triangleq \lim_{{K \to \infty}} \frac{1}{KT} \sum_{k=0}^{K-1} \sum_{t \in \mathcal{T}_k} \mathbb{E} \left\{E_n^t\right\} \leq \overline{E}_n, \label{constraint p2b}\\
        & 0 \leq \gamma_{n}^{k} \leq 1, k = 0,1,\cdots \label{constraint p2c}\\ 
        & 0 \leq p_{n}^{t} \leq \overline{P}_n, t = 0,1,\cdots \label{constraint p2d}
    \end{align}
\end{subequations}\par
To apply the Lyapunov optimization technique, we first convert the reliability constraint (\ref{constraint p2b}) into an equivalent virtual queue stability constraint. This is achieved by constructing an energy consumption deficit queue to measure the deviation between $E_n^t$ and $\overline{E}_n$, with the queue length evolving as  
\begin{align}\label{queue update}
    Q_n^{t+1} = \left[Q_n^t+E_n^t-\overline{E}_n\right]^{+}, \forall n \in \mathcal{N}, t = 0, 1, \cdots
\end{align}
where $[\cdot]^+ \triangleq \max \{\cdot, 0\}$, and $Q_n^0 = 0, \forall n \in \mathcal{N}$. For each device $n$ at slot $t$, the queue length $Q_n^t$ reflects how far the current energy consumption exceeds the budget $\overline{E}_n$. According to the Lyapunov optimization theory \cite{neely2010stochastic}, the long-term time-averaged constraint (\ref{constraint p2b}) is equivalent to the mean-rate stability constraint on the virtual queue, i.e., $\lim\limits_{t \to \infty} \mathbb{E}\{Q_n^t\}/t \to 0, \forall n \in \mathcal{N}$.\par
Then the conditional Lyapunov drift for each device $n$ can be written as
\begin{align}\label{Lyapunov drift}
    \Delta_{n,T}(Q_n^t) \triangleq \mathbb{E}\left\{ \frac{1}{2} \left({Q_n^{t+T}}\right)^2 - \frac{1}{2}\left({Q_n^t}\right)^2 \Big| Q_n^t\right\}, 
\end{align}
which measures the expected change in the quadratic function of the queue length after $T$ consecutive time slots. Intuitively, by minimizing $\Delta_{n,T}(Q_n^t)$, we can prevent the queue length from unbounded growth, and thus stabilize the queue $Q_n^t$. \par
According to the drift-plus-penalty algorithm of the Lyapunov optimization framework, the convergence error (as a penalty function) is incorporated into (\ref{Lyapunov drift}) to derive the following drift-plus-penalty function for the $k$-th frame of device $n$:
\begin{align}\label{drift-plus-penalty}
    \mathcal{D}_n(Q_n^{kT}) \triangleq \Delta_{n,T}(Q_n^{kT}) + V \mathbb{E}\left\{ \sum_{t \in \mathcal{T}_k} X_n^t \Big| Q_n^{kT} \right\},  
\end{align}
where $V \geq 0$ is a control parameter that indicates the emphasis we place on minimizing the convergence error. \par
The main idea of the Lyapunov optimization-based algorithm is to minimize the upper bound of the drift-plus-penalty term $\mathcal{D}_n(Q_n^{kT})$ to jointly guarantee the convergence error minimization and the energy consumption stability. To this end, we introduce the following two lemmas regarding the upper bound of $\mathcal{D}_n(Q_n^{kT})$ for our two-timescale algorithm design. 
\begin{lemma}\label{lemma:drift-plus-penalty upper bound 1}
    For each device $n \in \mathcal{N}$, under any feasible decisions $0\leq \gamma_{n}^{k} \leq 1$ and $0\leq p_n^t \leq \overline{P}_n, \forall t \in \mathcal{T}_k$, $\mathcal{D}_n(Q_n^{kT})$ is bounded by 
    \begin{align}\label{drift-plus-penalty 1}
    \mathcal{D}_n(Q_n^{kT}) &\leq \beta_1 T+\mathbb{E}\left\{ \sum_{t\in \mathcal{T}_k} Q_n^t(E_n^t-\overline{E}_n) \Big| Q_n^{kT}\right\} \nonumber \\
    &+ V \mathbb{E}\left\{ \sum_{t \in \mathcal{T}_k} X_n^t \Big| Q_n^{kT} \right\}, 
    \end{align}
where $\beta_1 = \frac{1}{2}(E_{n, \max}^2+\overline{E}_n^2)$ is a constant. 
\end{lemma}
\begin{proof}
    See Appendix C. 
\end{proof}
Minimizing the upper bound presented in Lemma \ref{lemma:drift-plus-penalty upper bound 1} is straightforward for single timescale (i.e., $T=1$). However, the decision variables of Problem $\mathcal{P}2$ should be iteratively adjusted at two different timescales. Applying this directly to the two-timescale case is challenging because minimizing the R.H.S. of (\ref{drift-plus-penalty 1}) at the beginning of every frame $t = kT$ depends on the future information of $\{Q_n^t\}$ over $t \in (kT +1, ...,(k+1)T-1)$, which is difficult to predict in practice due to its accumulative nature across time slots. To address this issue, we further relax the R.H.S. of (\ref{drift-plus-penalty 1}) as shown in the following lemma.
\begin{lemma}\label{lemma:drift-plus-penalty upper bound 2}
    For each device $n \in \mathcal{N}$, under any feasible decisions $0\leq \gamma_{n}^{k} \leq 1$ and $0\leq p_n^t \leq \overline{P}_n, \forall t \in \mathcal{T}_k$, we have 
    \begin{align}\label{drift-plus-penalty 2}
    &\mathcal{D}_n(Q_n^{kT}) \nonumber \\ &\leq \beta_2 T + \mathbb{E}\left\{ \sum_{t \in \mathcal{T}_k} VX_n^t+ Q_n^{kT}(E_n^t-\overline{E}_n) \Big| Q_n^{kT}\right\},
    \end{align}
    where $\beta_2 = \beta_1 + \frac{(T-1)[(E_{n, \max}-\overline{E}_n)E_{n, \max} + \overline{E}_n^2]}{2}$ is a constant. 
\end{lemma}
\begin{proof}
    See Appendix D. 
\end{proof}\par
The upper bound in Lemma \ref{lemma:drift-plus-penalty upper bound 2} is derived from the R.H.S. of (\ref{drift-plus-penalty 1}) by approximating the future queue length values as the current value at slot $kT$, i.e., $Q_n^t \approx Q_n^{kT}, \forall n \in \mathcal{N}$ for all $t \in \mathcal{T}_k$. This approximation avoids the prediction of future queue lengths, which significantly reduces the complexity and suits more on the two-timescale design. 
Then the joint problem with respect to $\gamma_n^k$ and $p_n^t$ for Lyapunov optimization is formulated as follows.\par
$\noindent\mathcal{P}3:$
\begin{subequations}
\begin{align}
\min_{\{p_{n}^{t}, \gamma_{n}^{k}\}} \quad & \sum_{k=0}^{K-1}\mathbb{E}\left\{ \sum_{t \in \mathcal{T}_k} VX_n^t+ Q_n^{kT}(E_n^t-\overline{E}_n) \Big| Q_n^{kT}\right\}\\
\text{s.t.} \quad & 0 \leq \gamma_{n}^{k} \leq 1, \forall  t \in \mathcal{T}_k, k = 1, 2, \cdots\\ 
& 0 \leq p_{n}^{t} \leq \overline{P}_n, \forall  t \in \mathcal{T}_k, k = 1, 2, \cdots
\end{align}
\end{subequations} \par
Note that given the current energy consumption deficit queue $Q_n^{kT}$, $Q_n^{kT}\overline{E}_n$ can be treated as a constant term and omitted from the Problem $\mathcal{P}3$. 
\subsection{Algorithm Design}
We now introduce the design of the online two-timescale algorithm. This algorithm aims to minimize the drift-plus-penalty upper bound, i.e., the second term on the R.H.S. of (\ref{drift-plus-penalty 2}), subject to the constraints (\ref{constraint p2c}) and (\ref{constraint p2d}), which can be proved to achieve a good performance for the original Problem $\mathcal{P}1$. Specifically, the algorithm operates in an online manner and takes the following three control actions:
\begin{itemize}
    \item (\textbf{Parameter freezing decision per frame}) At time slot $t = kT$, with $k = 0, 1, ...$,  given the current energy consumption deficit queue $Q_n^{kT}$, each device $n$ decides the optimal parameter freezing percentage $\gamma_{n}^{k}$ for the current frame by solving the following per-frame problem:
\begin{subequations}
\begin{align}\label{per frame problem}
      \mathcal{P}4: \quad \min_{ \{\gamma_{n}^{k}\}} \quad & \mathbb{E}\left\{ \sum_{t \in \mathcal{T}_k} VX_n^t+Q_n^{kT}E_n^t \right\}\\
\text{s.t.} \quad & 0 \leq \gamma_{n}^{k} \leq 1, \forall  t \in \mathcal{T}_k, \\ 
& 0 \leq p_{n}^{t} \leq \overline{P}_n, \forall  t \in \mathcal{T}_k, 
\end{align}
\end{subequations}
where the expectation $\mathbb{E}\{\cdot\}$ here is taken over the channel randomness $\{h_n^t, \forall n\}$, for all $\forall  t \in \mathcal{T}_k$. 
\item (\textbf{Power control decision per slot}) At every slot $t \in \mathcal{T}_k$, given the parameter freezing percentage $\gamma_{n}^{k}$, each device $n$ monitors the real-time channel condition $h_n^t$, and decides the transmit power $p_{n}^{t}$ by solving the following per-slot problem:
\begin{subequations}
\begin{align}\label{per slot problem}
      \mathcal{P}5: \quad \min_{\{p_{n}^{t}\}} \quad & VX_n^t+Q_n^{kT}E_n^t \\
\text{s.t.} \quad  & 0 \leq p_{n}^{t} \leq \overline{P}_n, \forall  t \in \mathcal{T}_k. 
\end{align}
\end{subequations}
\item (\textbf{Queue update}) At each slot  $t \in \mathcal{T}_k$, based on the obtained $p_n^{t,*}$, each device $n$ computes $E_n^t$ as (\ref{client energy consumtion}), and then updates the virtual queue $Q_n^t$ according to (\ref{queue update}).
\end{itemize}\par
We next develop the optimal solutions for the two different timescale problems, Problem $\mathcal{P}4$ and Problem $\mathcal{P}5$, respectively, which are highly non-trivial for the algorithm implementation.

\subsection{Algorithm Implementation}
We derive the optimal power control strategy and the optimal parameter freezing percentage for solving the per-slot Problem $\mathcal{P}5$ and the per-frame Problem $\mathcal{P}4$, respectively. 
\subsubsection{Power Control Decision Per Slot}
From (\ref{per slot problem}), assuming the learning delay requirement is met, each device $n$ can either choose to reduce the convergence error $X_n^t$ by consuming an amount of energy $E_n^t$ to upload local gradient parameters, or opt not to upload these parameters at the expense of $VX_n^t$. Moreover, the virtual queue length $Q_n^{kT}$ acts as the price for successfully updating and uploading local gradient parameters. A higher $Q_n^{kT}$ emphasizes more on energy consumption reliability, suggesting that devices should prioritize energy management to participate in as many training rounds as possible; while a lower $Q_n^{kT}$ indicates a preference for improving global model performance, tolerating more energy expenditure in each slot. Intuitively, as the queue evolves, the coordination between convergence error and energy consumption can be adaptively managed over frames. Next, we specify the optimal power strategy for Problem $\mathcal{P}5$ as follows.
\begin{proposition} \label{proposition 1}
The optimal transmit power for Problem $\mathcal{P}5$ is given by
\begin{align}\label{optimal power}
    p_n^{t, *} = 
    \begin{cases}
        p_{n,\min}^t, & \text{if $p_{n, \min}^t \leq p_{n, \max}^k$,} \\
        0, &\text{otherwise,}
    \end{cases}
\end{align}
where $p_{n, \min}^t$ represents the minimum transmit power required at slot $t$ to satisfy the learning latency requirement, while $p_{n, \max}^k$ denotes the maximum allowable power for per-slot local gradient uploading during the $k$-th frame, which are respectively defined as:
\begin{align}\label{min transmit power}
    p_{n}^{\min}(t) \triangleq \frac{N_0}{h_n^t} \left(2^{\frac{(1-\gamma_{n}^{k}) S }{W(\tau_0 - \tau_n^{{\rm cmp},t})}}-1\right),
\end{align}
\begin{align}\label{max transmit power}
    p_{n, \max}^k \triangleq \left[\frac{(V B_n \lambda - e_n^{{\rm cmp}}Q_n^{kT}) (1 - \gamma_{n}^{k})}{Q_n^{kT}(\tau_0  - \tau_n^{{\rm cmp},k})} \right]_{0}^{\overline{P}_n}, 
\end{align}
where $[\cdot]_{a}^{b} = \min \{ \max \{ \cdot, a\}, b\}$, and $e_n^{{\rm cmp}}=\frac{\alpha_n c_n B_n f_n^2}{2}$. 
\end{proposition}
\begin{proof}
It can be verified from (\ref{eqn:com_delay}) and (\ref{eqn:com_energy}) that $\tau_n^{{\rm com},t}$ is monotonically decreasing while $E_n^{{\rm com},t}$ is monotonically increasing with $p_n^t$, $\forall n \in \mathcal{N}$. Let $\tau_n^{{\rm com},t} + \tau_n^{{\rm cmp},t}= \tau_0$, we derive $p_{n, \min}^t$ in (\ref{min transmit power}) as the minimum required power to satisfy the learning latency constraint at slot $t$. Furthermore, setting $p_n^{t, *} = p_{n, \min}^t$ ensures the minimum energy consumption for device $n$ when uploading local gradient parameters at slot $t$.\par
We also observe that if the minimum energy consumption required for a device to participate in training exceeds the transmission outage cost, it becomes more energy-efficient for the device to quit the training, thereby setting $p_n^{t, *} = 0$. Additionally, we define the learning performance improvement when device $n$ participates in training as $ \Delta X_n^t = X_n(\mathbbm{1}_n^t = 1)-X_n(\mathbbm{1}_n^t = 0)$. By setting $V |\Delta X_n^t| = Q_n^{kT}E_n^t$, and incorporating the peak power constraint (\ref{constraint p2d}), we can derive $p_{n, \max}^k$ in (\ref{max transmit power}) and the condition $p_{n, \min}^t  \leq p_{n, \max}^k$ in (\ref{optimal power}), which completes the proof. 
\end{proof}
Proposition \ref{proposition 1} reveals that the optimal power control strategy for Problem $\mathcal{P}5$ follows a threshold-based policy. When $p_{n, \min}^t$ is below the threshold $p_{n, \max}^k$, the device uploads the local gradient in power $p_{n, \min}^t$; otherwise, the device should drop out of training (i.e., $p_n^{t, *}=0$) to avoid excessive energy consumption. Moreover, as the virtual queue length $Q_n^{kT}$ increases, the threshold $p_{n, \max}^k$ is further limited to reduce energy consumption, and thus ensure the stability of the queue. 
It is worth noting that $p_{n, \min}^t$ in (\ref{min transmit power}) changes over each slot, adapting to the real-time channel condition $h_n^t$, while $p_{n, \max}^k$ in (\ref{max transmit power}) remains unchanged within a frame but it is adjusted from one frame to another according to the updated $Q_n^{kT}$. 
\subsubsection{Parameter Freezing Decision Per Frame}
The optimal $\gamma_{n}^{k}$ can be obtained by solving the per-frame Problem $\mathcal{P}4$. As Problem $\mathcal{P}4$ is an expectation minimization problem, we compute the expectation by assuming that the channel randomness is i.i.d. over the slots of a frame, and each device has the statistical knowledge of channels in the current frame, but not the future frames. \par
% According to the optimal power control strategy in Proposition \ref{proposition 1}, 
Denote $z_n(\gamma_n^k) = VX_n^t+Q_n^{kT}E_n^t$ as the execution cost in slot $t$. Incorporating Proposition \ref{proposition 1}, we derive the expected optimal per-frame performance for Problem $\mathcal{P}4$ as follows. 
\begin{theorem}
Suppose that $h_n^t$ is i.i.d. over the slots of a frame with the probability density function (PDF) denoted by $f_{h_n^k}
$. Then, for all $t \in \mathcal{T}_k$, the expectation of $z_n(\gamma_n^k)$ taken over the channel randomness $h_n^t$ is obtained as
\begin{align}\label{expected cost 1}
    Z_n(\gamma_{n}^{k}) = \mathbb{E}\{ z_n(\gamma_n^k)\} & = VB_n \lambda( \gamma_n^k -1) \text{$\rm{Pr}$}\{h_n^t \geq h_{n, \min}^k\}\nonumber\\
    &\quad + Q_n^{kT} E_n^t \text{$\rm{Pr}$}\{h_n^t\geq h_{n, \min}^k\},
\end{align}
where $\rm{Pr}\{\cdot\}$ is the probability function. $h_{n, \min}^k$ is the minimum channel gain for device $n$ to upload local gradient successfully, which can be expressed as
\begin{align} \label{the minimum channel gain}
    h_{n, \min}^k \triangleq \frac{N_0}{p_{n, \max}^k} \left(2^{\frac{(1-\gamma_{n}^{k}) S }{W(\tau_0 - \tau_n^{{\rm cmp},k})}}-1\right). 
\end{align}
\end{theorem}
\begin{proof}
    According to Proposition \ref{proposition 1} and comparing $p_{n, \min}^t$ with $p_{n, \max}^k$, we can derive
    \begin{align}
        z_n(\gamma_n^k) = 
        \begin{cases}
            VB_n\lambda(\gamma_n^k-1)+Q_n^{kT}E_n^t,  & \text{if } h_n^t \geq h_{n, \min}^k; \\
            0,  & \text{otherwise}. 
        \end{cases}
    \end{align}
    Taking expectation on $z_n(\gamma_n^k)$ over the random variable $h_n^t$, we obtain $Z_n(\gamma_{n}^{k})$ in (\ref{expected cost 1}). Since $h_n^t$ follows the same distribution among slots $t \in \mathcal{T}_k$, $Z_n(\gamma_{n}^{k})$ is identical for every $t \in \mathcal{T}_k$, which completes the proof. 
\end{proof}
Note that $Z_n(\gamma_{n}^{k})$ in (\ref{expected cost 1}) represents the minimum expected execution cost (i.e., weighted sum of convergence error and energy consumption cost) for each slot $t \in \mathcal{T}_k$ under a stationary channel environment. It can be further expressed as
\begin{align} \label{expected cost 2}
    Z_n(\gamma_{n}^{k}) &= \left(VB_n \lambda - Q_n^{kT}e_n^{\rm{cmp}}\right)(\gamma_{n}^{k} -1) \int_{h_{n, \min}^k}^{+\infty} f_{h_n^k} dh \nonumber\\
    &\quad + Q_n^{kT} e_n^{{\rm com}, k} \int_{h_{n, \min}^k}^{+\infty}\frac{1}{h}f_{h_n^k} dh,
\end{align}
where 
\begin{align}
    e_n^{{\rm com},k} = N_0(\tau_0 - \tau_n^{{\rm cmp},k}) \left(2^{\frac{(1-\gamma_{n}^{k}) S }{W(\tau_0 - \tau_n^{{\rm cmp},k})}}-1\right). 
\end{align}
Here, $e_n^{{\rm com},k}$ and $h_{n, \min}^k$ in (\ref{expected cost 2}) are functions with respect to $\gamma_n^k$ for device $n$, while $Q_n^{kT}$ and $V$ (that affects $p_{n, \max}^k$) are known as constants at the beginning of the $k$-th frame. Therefore, with the statistical knowledge of channel conditions, the device $n$ is able to compute $Z_n(\gamma_{n}^{k})$ by (\ref{expected cost 2}) at the beginning of each frame $t = kT$. \par
It can be seen that the per frame Problem $\mathcal{P}4$ is a non-convex problem, due to the fact that the maximum transmit power $p_{n, \max}^k$ is a truncated function with respect to $\gamma_n^k$ according to (\ref{max transmit power}). This results in non-smoothness to the minimum channel gain $h_{n, \min}^k$ according to (\ref{the minimum channel gain}), making the objective function $Z_n(\gamma_{n}^{k})$ non-differentiable. To address this issue, we employ the majorization-minimization (MM) algorithm \cite{9151375} to solve the Problem $\mathcal{P}4$. In following, we consider the solution of Problem $\mathcal{P}4$ in each frame for each device, thus the superscript $k$ and the subscript $n$ are omitted for notational brevity. \par
Firstly, we initialize a feasible solution $\gamma^0$ for Problem $\mathcal{P}4$. Then in iteration $l+1, l=0, 1, 2 \cdots$, the surrogate function is constructed as follows: 
\begin{align}\label{surrogate function}
    \tilde{Z}(\gamma \mid \gamma^l) = {Z}(\gamma^l)+{Z}'(\gamma^l)(\gamma -\gamma^l)+\frac{M}{2}(\gamma-\gamma^l)^2,
\end{align}
where ${Z}'(\gamma^l)$ is the first-order derivative of ${Z}(\gamma)$ at the feasible point $\gamma^l$ and $M$ satisfies the inequality $M \geq Z''(\gamma), \forall \gamma \in [0, 1]$. Specifically, $M$ is given by

\begin{equation}\label{M value}
    M = \begin{cases}
    IG(1)Y(0), & \Gamma \leq 0, \gamma \in [0, 1]; \\
    \begin{aligned}
    e^{-\frac{h_{\min}(\Gamma)}{H}}\Big[&-(2I + P\overline{Q}\theta)\frac{h'_{\min}(0)}{H} \\
    &+ I\frac{h''_{\min}(0)}{H}\Big],
    \end{aligned}
    & \Gamma > 0, \gamma \in [0, \Gamma]; \\
    IG(1)Y(\Gamma), & \Gamma > 0, \gamma \in (\Gamma, 1].
    \end{cases}
\end{equation}
Here, $I = VB\lambda-Qe^{\rm{cmp}}$, $G(\gamma) = \frac{e^{-\frac{h_{\min}(\gamma)}{H}}}{h_{\min}(\gamma)}$, $Y(\gamma) = -2 h'_{\min}(\gamma) +(1-\gamma)h''_{\min}(\gamma)$, $\Gamma =1- \frac{\overline{P}Q\tau_0}{I+\overline{P}Q\theta}$ and $\theta = \frac{cB}{f}$. Then the following proposition can be established.\par
\begin{proposition} \label{proposition 2}
    The surrogate function $\tilde{Z}_n(\gamma \mid \gamma^l)$ satisfies the following conditions: 
    \begin{enumerate}
        \item Convexity condition: $\tilde{Z}(\gamma \mid \gamma^l)$ is a convex function with respect to $\gamma$;
        \item Local equality condition: $\tilde{Z}(\gamma^l \mid \gamma^l) = {Z}(\gamma^l)$, and $ \tilde{Z}'(\gamma^l \mid \gamma^l) = {Z}'(\gamma^l)$;
        \item Upper bound condition: $\tilde{Z}(\gamma \mid \gamma^l) \geq {Z}(\gamma)$. 
    \end{enumerate}
\end{proposition}
\begin{proof}
    See Appendix E. 
\end{proof}\par 
Secondly, we minimize the surrogate function (\ref{surrogate function}) to obtain the solution $\gamma^{\star}$. Let $\gamma^{l+1} = \gamma^{\star}$, we then proceed to construct the next-iteration surrogate function. \par
Thirdly, we iteratively perform the above two steps until convergence. Then we can obtain the optimal solution $\gamma^{*}$. \par
To better perform the MM-based optimal algorithm for problem $\mathcal{P}4$, we have the following lemma about the maximum transmit power $p_{\max}$ in (\ref{the minimum channel gain}) and the range of $\gamma$.
\begin{lemma}
\label{range of gamma}
The maximum transmit power $p_{\max}$ and the range of $\gamma$ satisfy the following properties:
\begin{enumerate}
    \item Case 1: if \( I \leq 0 \), then \( p_{\max} = 0 \), \(\forall \gamma \in [0, 1] \).
    \item Case 2: if \( 0 < I < \overline{P} Q(\tau_0 - \theta) \), then 
    \[ p_{\max} = \frac{I(1 - \gamma)}{Q(\tau_0 - \theta(1 - \gamma))} < \overline{P}_n, \gamma^{\min} = 0, \gamma^{\max} = 1. \]
    \item Case 3: if \( I \geq \overline{P} Q(\tau_0 - \theta) \), then we have:
    \begin{enumerate}
        \item if \( 0 \leq \gamma \leq \Gamma \), \( p_{\max} = \overline{P}, \gamma^{\min} = 0, \gamma^{\max} = \Gamma \),
        \item if \( \Gamma < \gamma \leq 1 \), 
        \[ p_{\max} = \frac{I(1 - \gamma)}{Q(\tau_0 - \theta(1 - \gamma))} < \overline{P}_n, \gamma^{\min} = \Gamma, \gamma^{\max} = 1. \]
    \end{enumerate}
\end{enumerate} 
\end{lemma}
\begin{proof}
    To prove Lemma \ref{range of gamma}, we consider the following three cases. \par
Case 1: According to (\ref{max transmit power}), a large $Q$ may result in $I = VB\lambda - Qe^{\rm{cmp}} \leq 0$. Then $p_{\max}$ is limited as zero in current frame, and thus the device quits participating in training. \par
Case 2: Let $\frac{I(1 - \gamma)}{Q(\tau_0  - \theta(1-\gamma))} < \overline{P}, \forall \gamma \in [0, 1]$, we can obtain that $\Gamma < \gamma \leq 1$. Then we can derive that
\begin{align}
    \begin{cases}
    I < \overline{P}Q(\tau_0 - \theta), &\Gamma <0;\\
    I \geq \overline{P}Q(\tau_0 - \theta), &\Gamma \geq 0.
\end{cases}
\end{align}
In this case, the maximum transmit power is further limited.\par 
Case 3: Let $\frac{I(1 - \gamma)}{Q(\tau_0  - \theta(1-\gamma))} \geq \overline{P}, \forall \gamma \in [0, 1]$, we can obtain that $0 \leq \gamma \leq \Gamma$ and $I \geq \overline{P}Q(\tau_0 - \theta)$. In this case, $p_{\max}$ is only limited by the peak power $\overline{P}$. Thus we complete the proof.
\end{proof}\par
Case 1 indicates that excessive energy is consumed in previous frames, resulting in a large energy queue length $Q$, thus the device quits participating in training for current frame to ensure the stability of energy consumption.
Cases 2 and 3(b) show that the device prefers to further limit its transmit power and freeze more parameters to reduce the current queue length. Case 3(a) suggests that when the current queue length is small, the device focuses more on improving the model performance by reducing the parameter freezing percentage and increasing the transmit power, rather than prioritizing the stability of energy consumption. \par In summary, we summarize the proposed online two-timescale algorithm in Algorithm \ref{Online Two-Timescale Algorithm}. 

\begin{algorithm}[t]
	\renewcommand{\algorithmicrequire}{\textbf{Input:}}
	\renewcommand{\algorithmicensure}{\textbf{Output:}}
	\caption{Online Two-Timescale Algorithm}
	\label{Online Two-Timescale Algorithm}
	\begin{algorithmic}[1]
        \STATE Set $V \geq 0$, $\overline{E}_n$, and $\overline{P}_n$. 
        \STATE Initialize $t=0$ and $Q_n^0 = 0$. 
        \FOR{each frame $k = 0, 1, \dots, K-1$}
        \STATE Set $\gamma_n^{k, \min}, \gamma_n^{k, \max}, \gamma_n^{k, 0}$.
        \REPEAT
        \STATE Obtain a feasible solution $\gamma_n^{k, \star}$ by minimizing (\ref{surrogate function});
        \STATE Update the starting point as $\gamma_n^{k, l+1} = \gamma_n^{k, \star}$;
        \UNTIL{convergence};
        \FOR{each slot $t = kT, kT + 1, \dots, (k+1)T - 1$}
        \STATE Compute $p_n^t, E_n^t$ by (\ref{optimal power}) and (\ref{client energy consumtion}), respectively.
        \STATE Update $Q_n^t$ by (\ref{queue update}).
        \ENDFOR
        \ENDFOR
        \STATE \textbf{Output:} $\{\gamma_n^k\}$ and $\{p_n^t\}$.
\end{algorithmic}  
\end{algorithm}
\setlength{\textfloatsep}{0.3cm}

\subsection{Performance analysis}
% \vspace{-0.1cm}
In this subsection, we present the performance bounds of the proposed algorithm. For ease of exposition, we assume that there exists a constant $\delta > 0$ and a feasible solution to Problem $\mathcal{P} 1$ so that the following inequality holds for all frames:
\begin{align}\label{average energy condition}
    \frac{1}{T}\mathbb{E}\{ \sum_{t \in \mathcal{T}_k} E_n^t \} < \overline{E}_n - \delta, \forall n \in \mathcal{N}. 
\end{align}\par
Then the performance bounds of the proposed algorithm can be described in the following theorem.
\begin{theorem}\label{algorithm performance analysis}
Assume that the condition (\ref{average energy condition}) is satisfied for $\exists \delta > 0$, and the initial virtual queue length is zero for device $n \in \mathcal{N}$, i.e., $Q_n^0 = 0, \forall n \in \mathcal{N}$. Then, for any $V > 0$, we have:
1) The average queue length under the proposed algorithm is upper bounded by
\begin{align} \label{upper bound of queue length}
    \lim_{K \to \infty} \frac{1}{KN} \sum_{k=0}^{K-1} \sum\limits_{n=1}^{N} \mathbb{E} \left\{ Q_{n}^{kT, *}\right\} \leq \frac{\beta_{2, \rm{av}} + VX_{\max, \rm{av}}}{\delta}, 
\end{align}
2) The average convergence error under the proposed algorithm is upper bounded by
\begin{align} \label{upper bound of convergence error}
    \lim_{K \to \infty} \frac{1}{KTN} \sum_{k=0}^{K-1}\sum_{n=1}^{N} \sum_{t \in \mathcal{T}_k} \mathbb{E} \left\{ X_n^{t, *}\right\} \leq \frac{\beta_{2, \rm{av}}}{V} + X_{\rm av}^{\rm opt}, 
\end{align}
where $\beta_{2,\rm{av}} = \frac{1}{N}\sum_{n=1}^{N} \beta_{2,n}$, $X_{\rm{av}, \max} = \frac{1}{N}\sum_{n=1}^{N} X_{n, \max}$, and $X_{\rm{av}}^{\rm{opt}} = \frac{1}{N}\sum_{n=1}^{N} X_{n}^{\rm{opt}}$.
\end{theorem}
\begin{proof}
    According to Lemma \ref{lemma:drift-plus-penalty upper bound 2} and the fact that the proposed algorithm is developed through minimizing the R.H.S. of the inequality (\ref{drift-plus-penalty 2}), $\forall n \in \mathcal{N}$, we have
\begin{align}
    &\mathcal{D}_n(Q_n^{kT, *})= \Delta_{n,T}(Q_n^{kT})+V\mathbb{E}\{ \sum_{t \in \mathcal{T}_{k}} X_n^{t, *} \Big| Q_n^{kT, *}\}\nonumber\\
    &\leq \beta_{2, n}T + \mathbb{E} \{ \sum_{t \in \mathcal{T}_{k}} VX_n^{t, *} + Q_n^{kT, *}(E_n^{t, *} - \overline{E}_n)|Q_n^{kT, *}\} \nonumber\\
    &\overset{(a)}{\leq} \beta_{2, n}T + \mathbb{E} \{ \sum_{t \in \mathcal{T}_{k}} V\widehat{X}_n^t + Q_n^{kT,*}(\widehat{E}_n^t - \overline{E}_n)|Q_n^{kT, *}\}\nonumber\\
    &\overset{(b)}{\leq} \beta_{2, n}T + \mathbb{E} \{ \sum_{t \in \mathcal{T}_{k}} V\widehat{X}_n^t\} - Q_n^{kT, *}\delta T.
\end{align}
Here, $\widehat{X}_n^t$ and $\widehat{E}_n^t$ denote the convergence error and energy consumption achieved by the policy
satisfying the condition (\ref{average energy condition}), respectively. Inequality $(a)$ and $(b)$ are due to the conditions (\ref{average energy condition}). Moreover, due to $\gamma_n^k \in [0, 1]$, we have
\begin{align}
    |X_n^{t, *}- \widehat{X}_n^t|= \lambda B_n |\gamma_n^{k, *} - \widehat{\gamma}_n^k| \leq \lambda B_n = X_{n, \max}.
\end{align}\par
Then we can obtain that 
\begin{align}
    &\frac{1}{2}\mathbb{E}\left\{ (Q_n^{kT,*})^2\right\} - \frac{1}{2}\mathbb{E}\left\{ (Q_n^{0, *})^2\right\}\nonumber\\
    &\leq K[\beta_{n, 2}T +  VTX_{n, \max}] - \delta T \sum_{k=0}^{K-1}\mathbb{E}\left\{ Q_n^{kT, *}\right\}.
\end{align}\par
Dividing both sides of $K\delta T$, taking limit as $ K \rightarrow \infty$ yield
\begin{align}\label{average single queue}
    &\lim_{K \rightarrow \infty} \frac{1}{K}\sum_{k=1}^{K-1} \mathbb{E}\left\{Q_n^{kT,*}\right\} \nonumber\\
    &\leq \frac{\beta_{2,n}+VX_{n, \max}}{\delta} + \frac{\mathbb{E}\left\{ (Q_n^{0,*})^2\right\} - \mathbb{E}\left\{ (Q_n^{kT, *})^2\right\}}{2K\delta T}\nonumber\\
    &\leq \frac{\beta_{2,n}+VX_{n, \max}}{\delta}.
\end{align}\par
By averaging over all devices, we prove (\ref{upper bound of queue length}). \par
According to \cite{neely2010stochastic}, if the problem is feasible, there exists a stationary optimal $\omega$-only policy, in which decisions $\{\gamma_n^k\}$ and $\{p_n^t\}$ are made independent of the queue length, achieving the minimum convergence error $X_{n,\rm{av}}^{\rm{opt}}$ while meeting the queue stability constraint. Therefore, $\forall n \in \mathcal{N}$, we have
\begin{align}\label{w-only policy}
    &\mathcal{D}_n(Q_n^{kT, *}) \leq \beta_{2, n}T + V\mathbb{E}\{\sum_{t \in \mathcal{T}_k} X_n^{t,{\rm{opt}}}\}. 
\end{align}
Taking expectation of (\ref{w-only policy}) and summing it over $k$ yield
\begin{align}
    &\frac{1}{2}\mathbb{E}\left\{ (Q_n^{kT, *})^2\right\} - \frac{1}{2}\mathbb{E}\left\{ (Q_n^{0, *})^2\right\} + V\sum_{k=1}^{K-1}\mathbb{E}\{ \sum_{t \in \mathcal{T}_k} X_n^{t, *}\}\nonumber\\
    &\leq KT\beta_{n, 2} +  V\sum_{k=1}^{K-1} \mathbb{E}\{\sum_{t \in \mathcal{T}_k}X_{n}^{t, {\rm{opt}}}\}.
\end{align}\par
Similar to (\ref{average single queue}), by dividing both sides of $KVT$, taking limit as $ K \rightarrow \infty$, and averaging over all devices, we prove (\ref{upper bound of convergence error}).   
\end{proof}
Theorem \ref{algorithm performance analysis} shows that the average convergence error of the proposed online algorithm can asymptotically achieve the optimum $X_{\rm{av}}^{\rm opt}$ of Problem $\mathcal{P}1$ by increasing the control parameter $V$. Moreover, the average virtual queue length is upper bounded by $\mathcal{O}(V)$ as shown in (\ref{upper bound of queue length}), indicating the queue is mean rate stable and the reliability constraint (\ref{constraint 1}) is guaranteed.

In terms of computational complexity of Algorithm \ref{Online Two-Timescale Algorithm}, with the stopping criterion for the MM algorithm set to 
$|\gamma^{l+1} - \gamma^{l}| \leq \epsilon$, the complexity of performing the MM algorithm in each frame is $\mathcal{O}(\log(\frac{1}{\epsilon
}))$ (corresponding to Steps 5 through 8 in Algorithm \ref{Online Two-Timescale Algorithm}) \cite{boyd2004convex}. Moreover, since each frame contains $T$ slots and each slot has a computational complexity of 
$\mathcal{O}(1)$, the overall computational complexity of Algorithm \ref{Online Two-Timescale Algorithm} is $\mathcal{O}(K\log(\frac{1}{\epsilon
})+KT)$, where $K$ is the total number of frames. 

\vspace{-0.1cm}
\section{Experimental Results}
% \vspace{-0.1cm}
\label{sec:Experimental_Results}
In this section, we conduct experiments on public datasets to evaluate the performance of the proposed two-timescale FL scheme. 
\subsection{Experiment Settings}
\textit{Wireless Network Setting}: We consider a wireless network in which the server and 30 devices are in a circle area with a radius of 1000 meters. We refer to a single communication round as a slot and group every consecutive 20 time slots as a frame. The channel gains are modeled as $h_n^t = g_n^tH_n^k$. Specifically, the large-scale fading $H_n^k$ is generated according to the path-loss model $PL$ [dB] $= 128.1 + 37.6 \log_{10} d_n^k$, where $d_n^k$ is the distance in meters between device $n$ and the server in the $k$-th frame; the small-scale fading $g_n^t$ follows normalized exponential distribution. Besides, the noise power $N_0$ is $-104$ dBm \cite{10346989}. Each device is allocated an equal bandwidth of 10 MHz. The CPU frequencies of all devices are set to 2 GHz.  As for the effective capacitance coefficient $\{\alpha_n\}$, similar to \cite{zeng2021energy}, we set $\alpha_n = 2 \times 10^{-28}$. Moreover, similar to \cite{9593178, 10089235}, we assign different energy budgets to devices by randomly sampling from a range of [0.30, 0.45] J for those training on the MNIST dataset and a range of [0.8, 1.0] J for those training on the CIFAR-10 dataset. \par

\textit{FL Setting}: We evaluate the proposed scheme under two widely used datasets: MNIST and CIFAR-10. More details of the experimental setup are presented as follows.\par
1) MNIST dataset: In the experiments, a 6-layer Convolutional Neural Network (CNN) model with 421642 parameters is trained. Specifically, the network consists of two convolutional layers, each followed by a ReLU activation and a 2 $\times$ 2 max pooling operation, and two fully connected (FC) layers. Both convolutional layers employ a 5 $\times$ 5 kernel with a stride of 1, with the first and second layers containing 20 and 50 filters, respectively. The extracted feature maps are flattened into a 4 $\times$ 4 $\times$ 50 vector and subsequently processed by two FC layers with 512 and 10 neurons, respectively. We use Dirichlet distribution Dir$(\rho)$ to generate both IID and non-IID data partitions among devices \cite{yurochkin2019bayesian}, where $\rho$ is the Dirichlet parameter. Specifically, we set $\rho=0.3$ for non-IID data partition and $\rho \to \infty$ for IID data partition. Besides, the learning rate is set at 0.05, the local training data size $B_n$ at 512 and the total latency in each slot $\tau_0$ at 800 ms \cite{9237168}.\par
2) CIFAR-10 dataset: We consider a MobileNetV2 model with 543050 parameters in the experiments \cite{sandler2018mobilenetv2}. Data partitioning still follows the Dirichlet distribution and $\rho = 0.3$ is set for non-IID data partition, $\rho \to \infty$ is set for IID data partition. The local training data size $B_n$ is also set at 512. Moreover, the learning rate is set as $0.05 \times {0.5^{\frac{t}{KT}}}$, where $KT$ is the total communication rounds. The total latency in each slot $\tau_0$ is set at 2 s \cite{9609994}. 
\vspace{-0.2cm}
\subsection{Performance Comparison with Heuristic Schemes}
\vspace{-0.1cm}
To demonstrate the effectiveness of the proposed scheme, we introduce several baseline schemes as follows. 
\begin{itemize}
    \item Ideal FL (Ideal): All local model parameters are not frozen. Moreover, all devices upload local gradient parameters without suffering transmission outages. \par
    \item Only power control (Only-PC): All local model parameters are not frozen (i.e., $\gamma_n^k =0$). And each device performs the power control strategy in Proposition \ref{proposition 1} when uploading the local gradient parameters. \par
    \item Only parameter freezing (Only-PF): Each device freezes the local model parameters as in the proposed scheme. Moreover, all devices upload local parameters without suffering transmission outages. 
\end{itemize}
\begin{figure}[t]
\centering
\subfigure[MNIST dataset.]{\label{fig:Impact of control parameter V MNIST}
    \includegraphics[clip, viewport= 5 5 710 560,width=0.45 \linewidth]{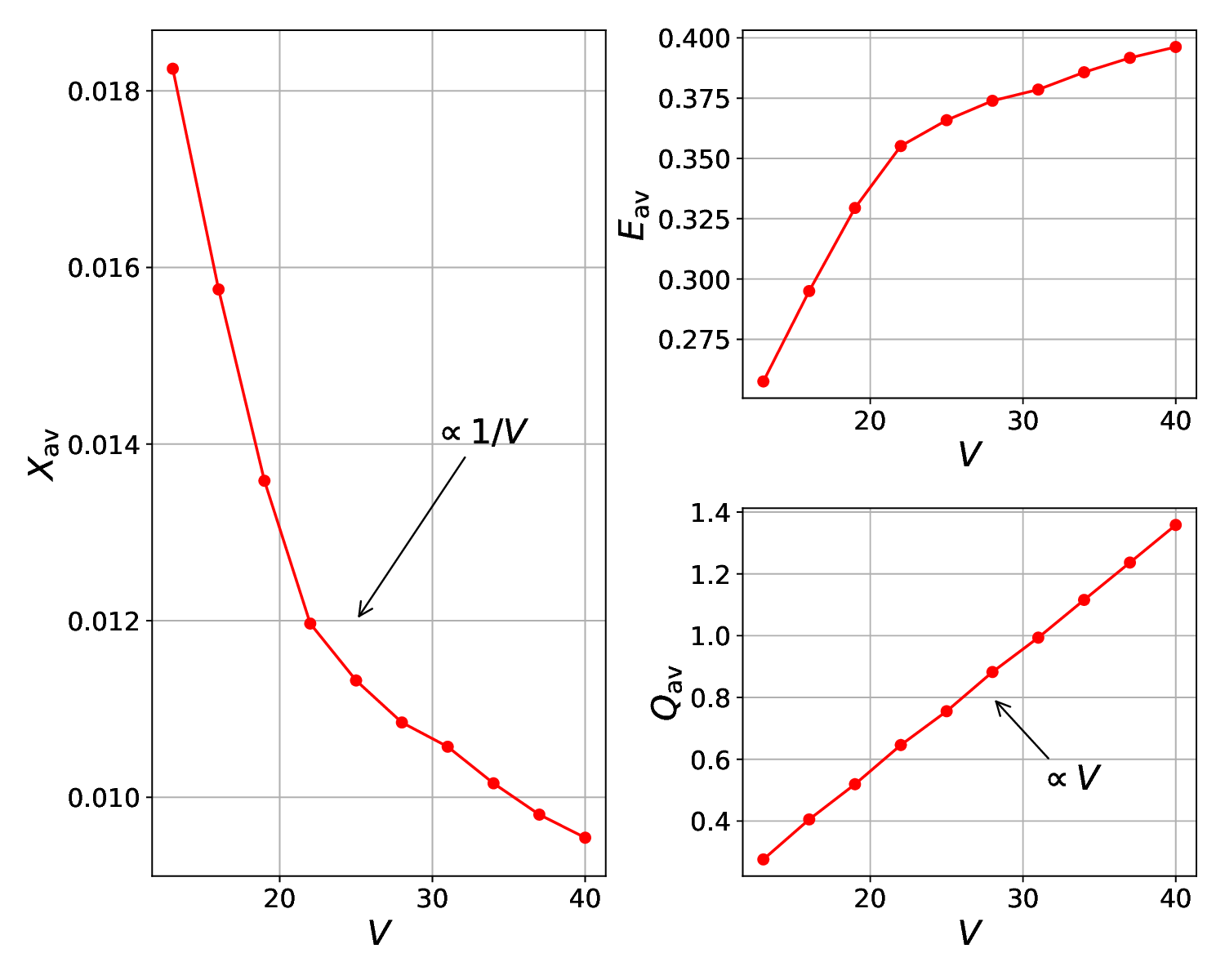}
}
\ 
\subfigure[CIFAR-10 dataset.]{\label{fig:Impact of control parameter V CIFAR10}
    \includegraphics[clip, viewport= 5 5 710 560,width=0.45 \linewidth]{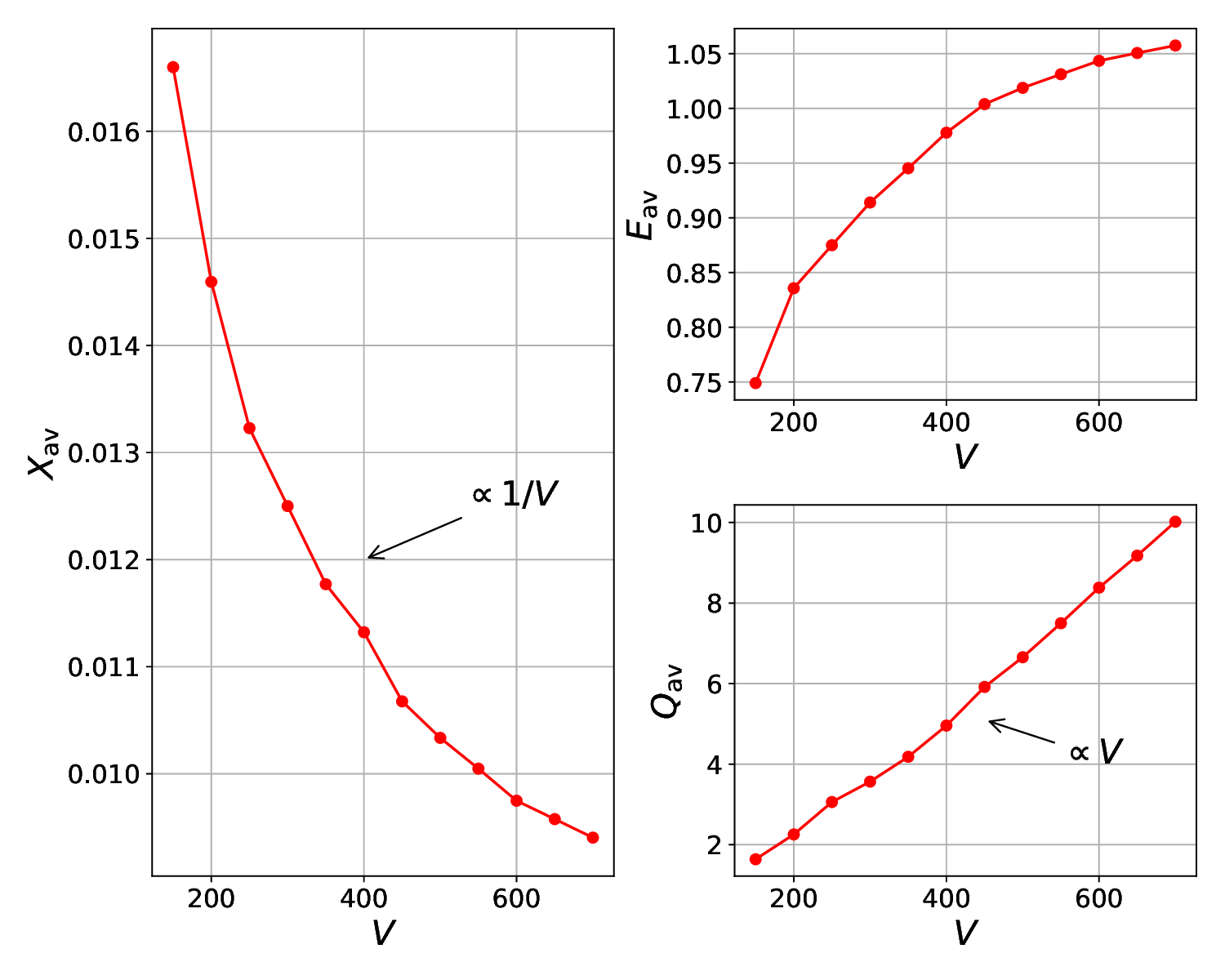}
}
% \vspace{-0.11in}
\caption{Impact of control parameter $V$ on (a) MNIST dataset and (b) CIFAR-10 dataset. }
\label{fig:Impact of control parameter V}
\end{figure}\par
Fig. \ref{fig:Impact of control parameter V} shows the impact of control parameter $V$ on the average convergence error, the average energy consumption, and the average virtual queue length of the proposed Algorithm \ref{Online Two-Timescale Algorithm}. We can observe that the average convergence error decreases inversely proportional to $V$ while the average energy consumption increases with $V$. This is because that a larger $V$ implies that the devices focus more on the improvement in model performance than the energy consumption, which forces the devices to freeze fewer parameters to minimize the convergence error and increase the transmit power to ensure reliable parameter transmission. Moreover, the average virtual queue length is increases linearly with $V$. These observations are consistent with the performance analysis of the proposed Algorithm \ref{Online Two-Timescale Algorithm} in Theorem \ref{algorithm performance analysis}. \par
\begin{figure}[t]
\centering
% \subfigcapskip=-5pt
\subfigure[Time evolution: average energy consumption on MNIST dataset.]{\label{fig:Time_evolution_average_energy_consumption_MNIST}
    \includegraphics[clip, viewport= 5 0 420 330,width=0.45 \linewidth]{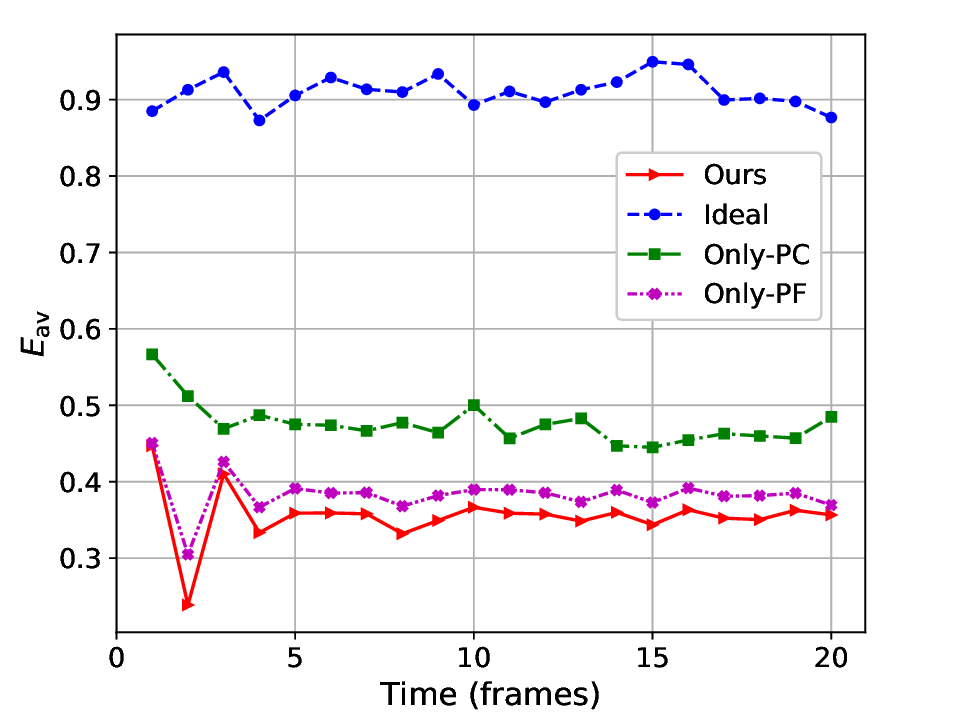}
}
\
\subfigure[Time evolution: average cost on MNIST dataset.]{\label{fig:Time_evolution_average_cost_MNIST}
    \includegraphics[clip, viewport= 5 0 420 330,width=0.45 \linewidth]{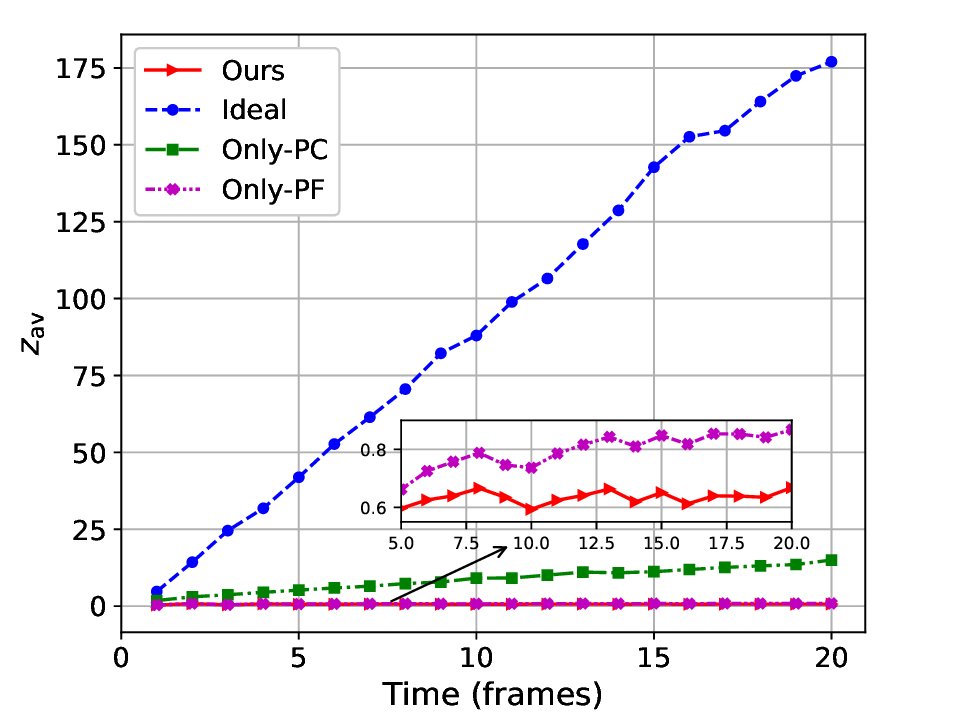}
}
% \vspace{-0.11in}
\caption{Time evolution: (a) average energy consumption and (b) average cost on MNIST dataset.}
\label{fig: Time evolution on MNIST}
\end{figure}
\begin{figure}[t]
\centering
% \subfigcapskip=-5pt
\subfigure[Time evolution: average energy consumption on CIFAR-10 dataset.]{\label{fig:Time_evolution_average_energy_consumption_CIFAR10}
    \includegraphics[clip, viewport= 5 0 420 330,width=0.45 \linewidth]{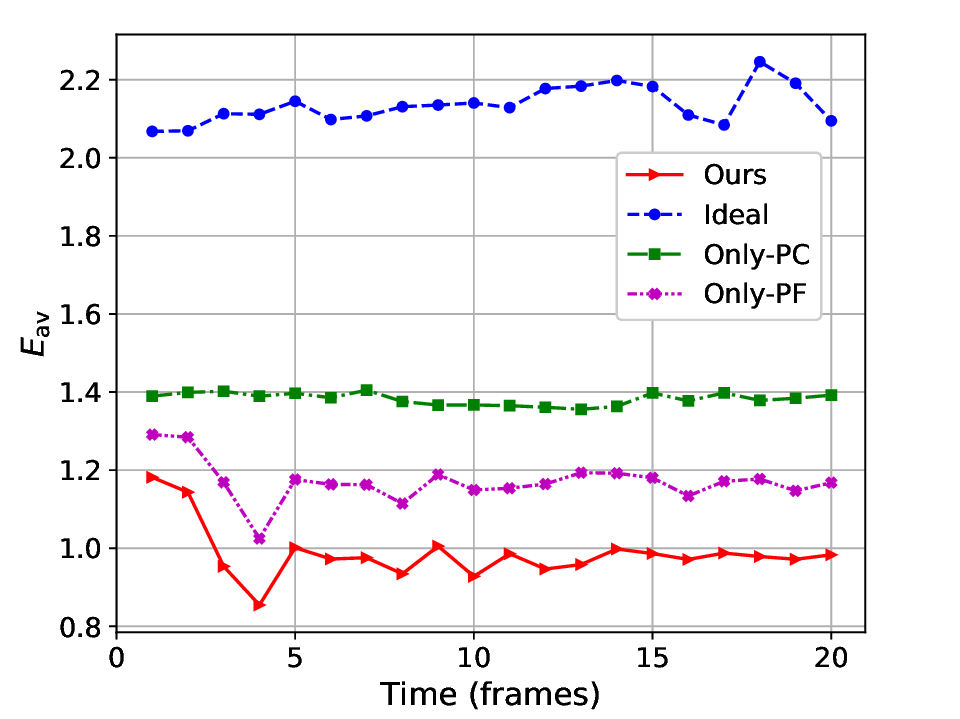}
}
\ 
\subfigure[Time evolution: average cost on CIFAR-10 dataset.]{\label{fig:Time_evolution_average_cost_CIFAR10}
    \includegraphics[clip, viewport= 5 0 420 330,width=0.45 \linewidth]{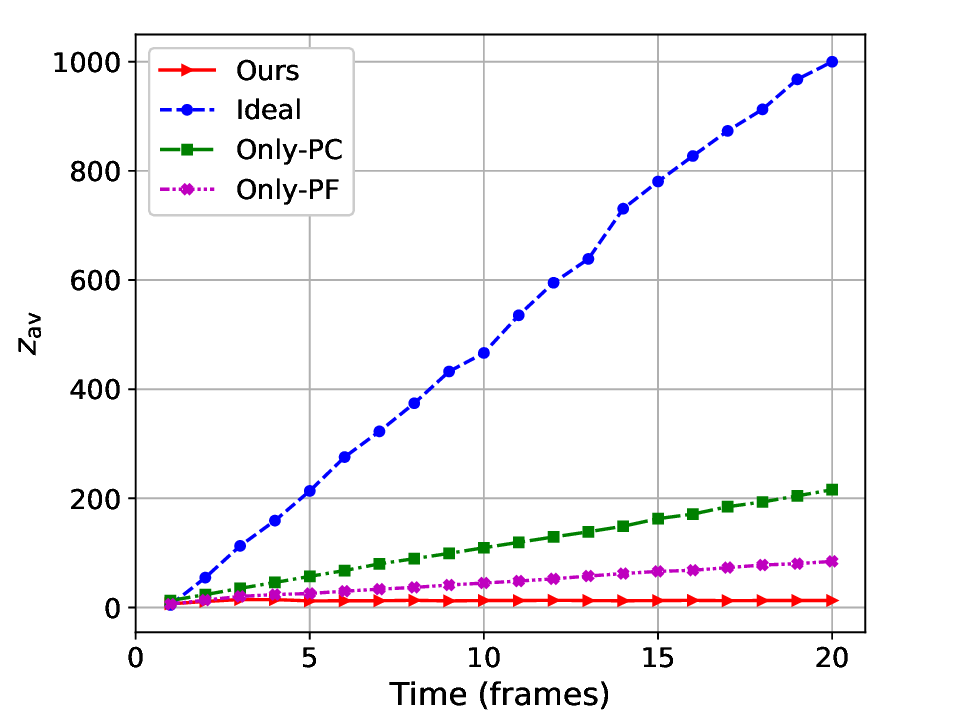}
}
% \vspace{-0.11in}
\caption{Time evolution: (a) average energy consumption and (b) average cost on CIFAR-10 dataset. }
\label{fig: Time evolution on CIFAR-10}
\end{figure}

Fig. \ref{fig: Time evolution on MNIST} and Fig. \ref{fig: Time evolution on CIFAR-10} show the average energy consumption $E_{\rm{av}}$ and the average cost $z_{\rm{av}}$ of all schemes over 20 frames on MNIST dataset and CIFAR-10 dataset respectively. We can observe that the proposed scheme consumes less energy than the three benchmarks and achieves the minimum average cost. The reason is that the proposed scheme can adaptively adjust the parameter freezing percentage and the transmit power during the training process to improve energy efficiency and simultaneously guarantee the learning performance. Among the benchmarks, we can see that the Ideal FL scheme incurs the highest energy consumption, resulting in the maximum average cost. This is because each device needs to consume excessive energy to update and upload the entire model gradient and ensure the success of parameter transmission, even under poor channel condition, which leads to a larger deficit queue length of energy consumption than other schemes. The Only-PC scheme can reduce excessive energy consumption but results in a high convergence error, and thus achieves a higher cost. The reason is that many devices consume excessive energy with small improvements in model performance, thus choosing not to participate in training. Conversely, the Only-PF scheme has a smaller convergence error but at expense of high energy consumption to ensure all devices participate in training. Moreover, we can also observe that the average energy consumption of the proposed scheme tends to stabilize over frames. This is because the proposed scheme can dynamically manage the parameter freeze rate and transmit power to achieve a well balance between the learning performance and energy consumption.

\begin{figure}[t]
\setlength{\belowcaptionskip}{-0.3cm}
\centering
% \subfigcapskip=-5pt
\subfigure[MNIST dataset (non-IID).]{\label{fig:Test accuracy vs. total energy consumption MNIST non IID}
    \includegraphics[clip, viewport= 3 3 432 335,width=0.45 \linewidth]{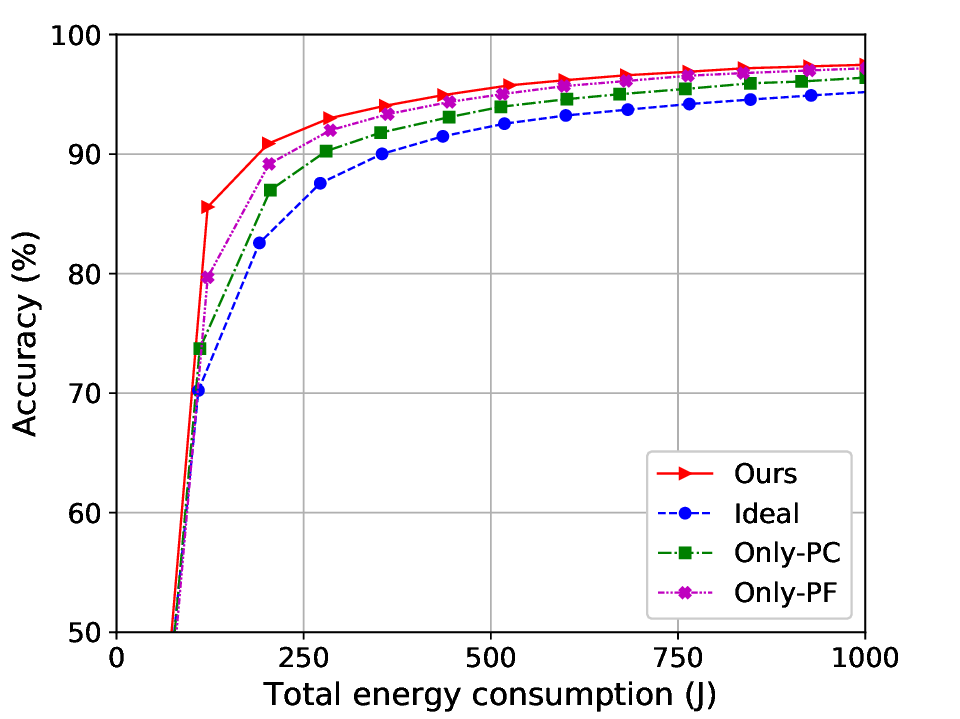}
}
\ 
\subfigure[CIFAR-10 dataset (non-IID).]{\label{fig:Test accuracy vs. total energy consumption CIFAR10 non IID}
    \includegraphics[clip, viewport= 3 3 432 335,width=0.45 \linewidth]{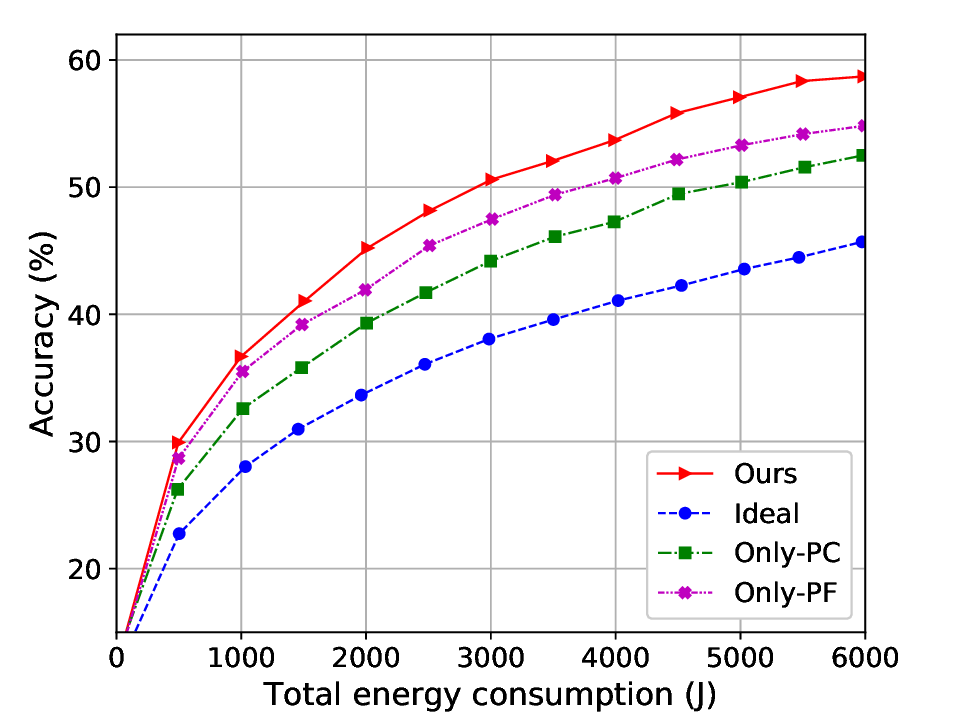}
}
\ 
\subfigure[MNIST dataset (IID).]{\label{fig:Test accuracy vs. total energy consumption MNIST IID}
    \includegraphics[clip, viewport= 3 3 432 335,width=0.45 \linewidth]{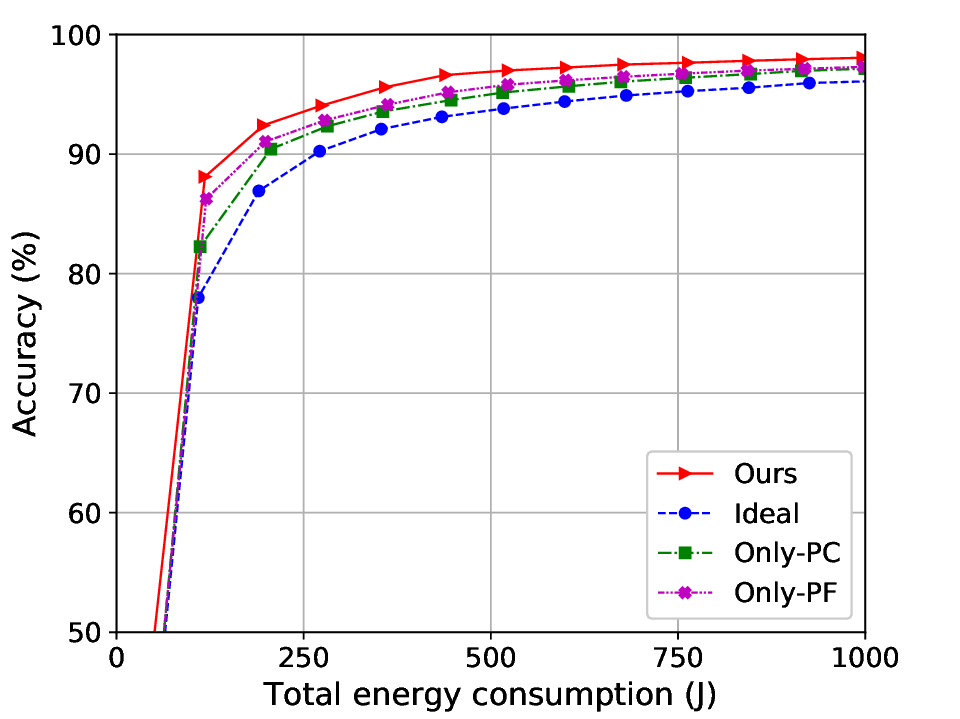}
}
\ 
\subfigure[CIFAR-10 dataset (IID).]{\label{fig:Test accuracy vs. total energy consumption CIFAR10 IID}
    \includegraphics[clip, viewport= 3 3 432 335,width=0.45 \linewidth]{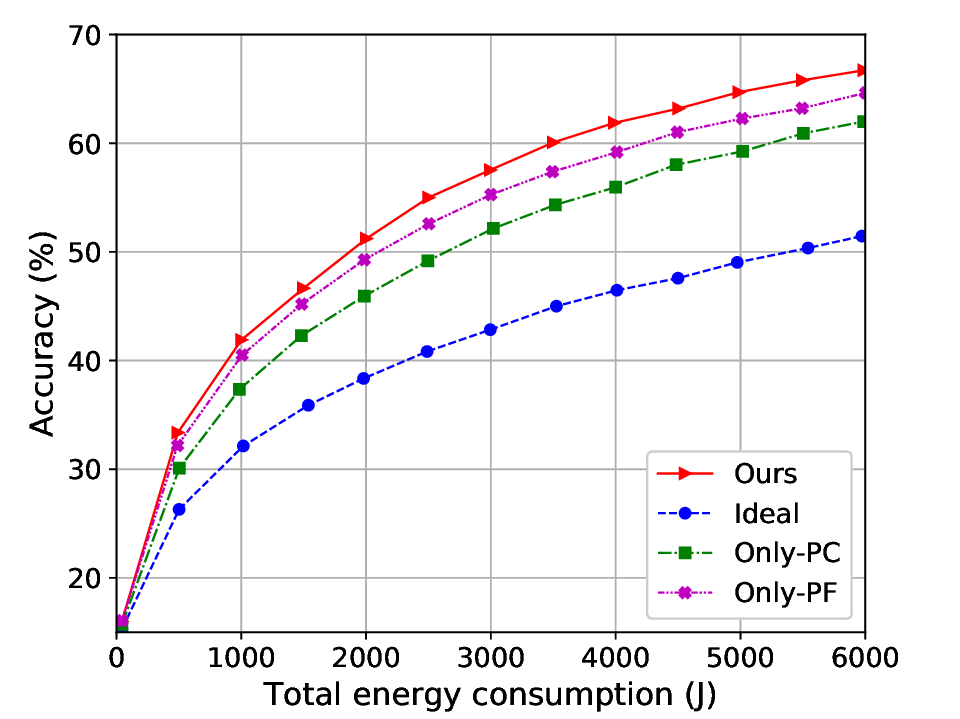}
}
% \vspace{-0.11in}
\caption{Test accuracy vs. total energy consumption on (a), (c) MNIST dataset and (b), (d) CIFAR-10 dataset with IID and non-IID data. }
\label{fig:Test accuracy vs. total energy consumption}
\end{figure}

Fig. \ref{fig:Test accuracy vs. total energy consumption} shows the test accuracy of all schemes versus the total energy consumption of the devices on non-IID and IID data. As expected, the proposed scheme can achieve higher accuracy than the three benchmarks at the same total energy consumption level. This is because the proposed scheme freezes the stable parameters from updating and uploading to save unnecessary energy consumption and adaptively adjusts the transmit power of devices to improve energy efficiency. Notably, the proposed scheme outperforms the Ideal FL scheme as it reduces the model dimensionality and thus boosts the convergence. 

\begin{figure}[t]
\centering
\subfigure[CIFAR-10 dataset (non-IID).]{
    \includegraphics[clip, viewport= 3 3 432 335,width=0.45 \linewidth]{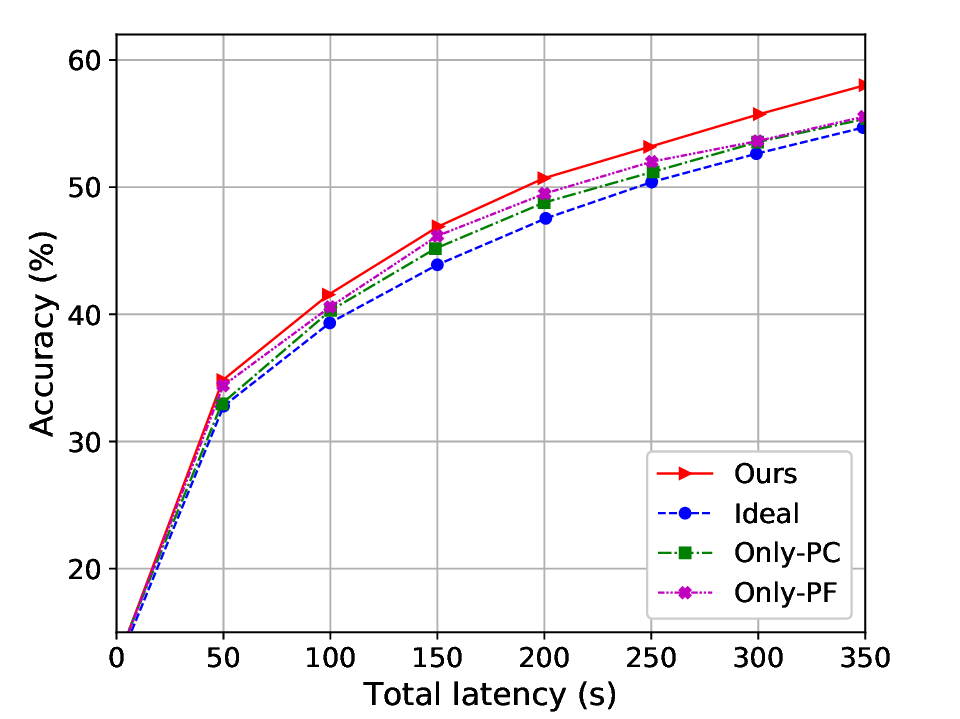}
}
\ 
\subfigure[CIFAR-10 dataset (IID).]{
    \includegraphics[clip, viewport= 3 3 432 335,width=0.45 \linewidth]{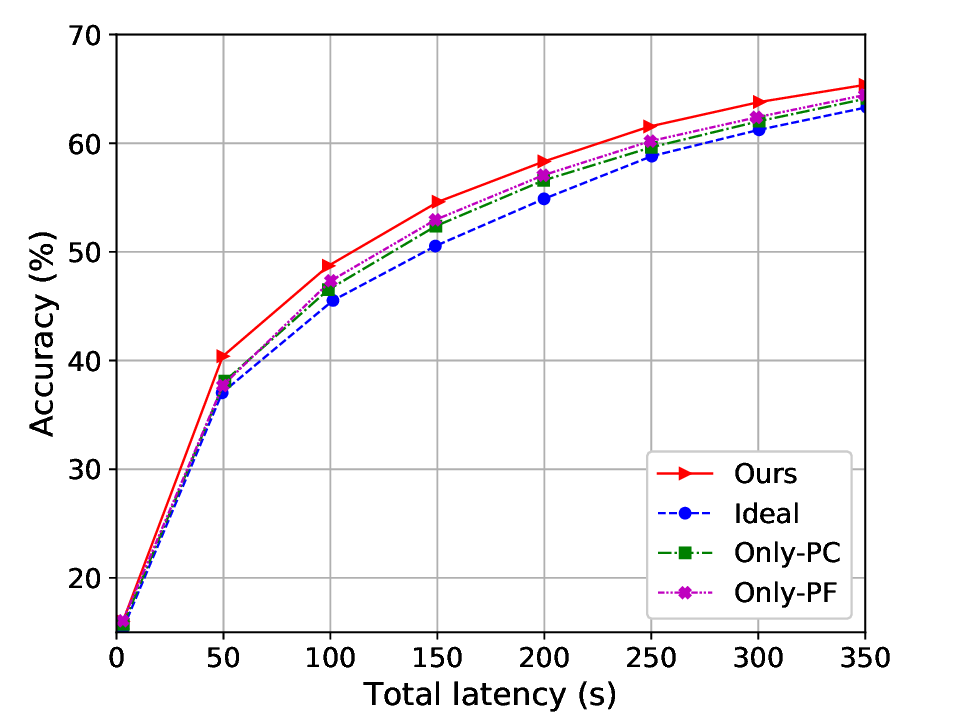}
}
% \vspace{-0.1in}
\caption{Test accuracy vs. total latency on CIFAR-10 dataset with IID and non-IID data. }
\label{fig:Test accuracy vs. total latency consumption}
\end{figure}\par
Fig. \ref{fig:Test accuracy vs. total latency consumption} shows the test accuracy of all heuristic schemes versus the total latency on non-IID and IID data. We observe that the proposed scheme outperforms both the Ideal FL and Only-PF schemes. This is because the Ideal FL and Only-PF schemes require additional communication latency to prevent transmission outages, even under peak transmit power. Moreover, the requirement to transmit the entire set of model parameters increases the communication load across multiple devices, resulting in a higher likelihood of transmission outages, which ultimately degrades the performance of the Only-PC scheme.

\begin{figure}[t]
\centering
% \subfigcapskip=-10pt
\subfigure[Average energy consumption vs. $\overline{E}_n$ on MNIST dataset.]{\label{fig:Average energy consumption vs E MNIST}
    \includegraphics[clip, viewport= 5 0 420 330,width=0.45\linewidth]{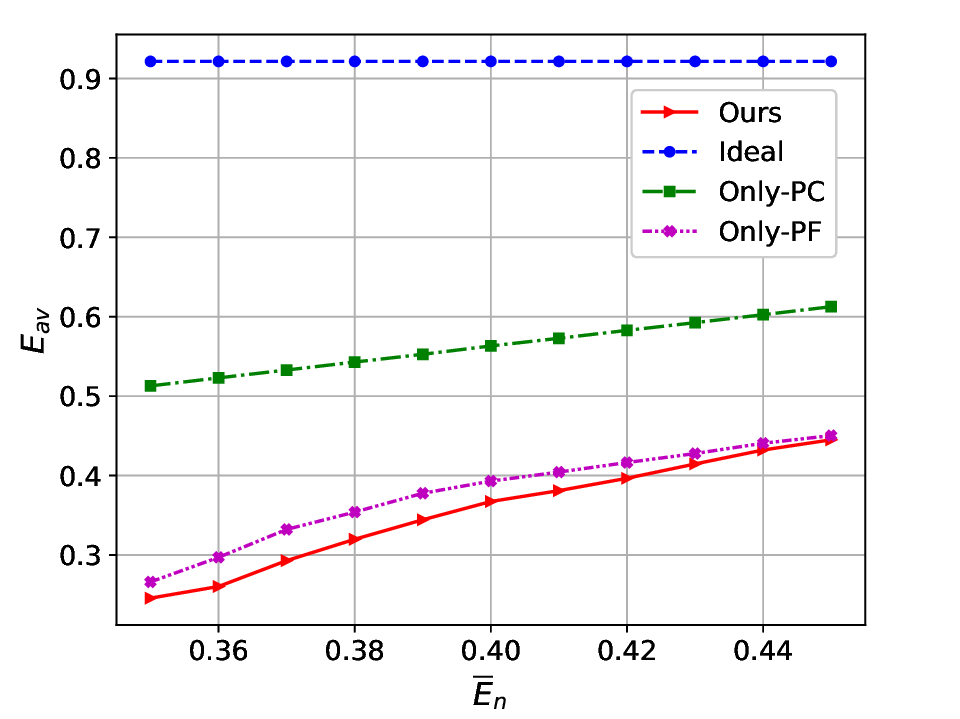}}
\ 
\subfigure[Average queue length vs. $\overline{E}_n$ on MNIST dataset.]{\label{fig: Average convergence error vs E MNIST}
    \includegraphics[clip, viewport= 5 0 420 330,width=0.45 \linewidth]{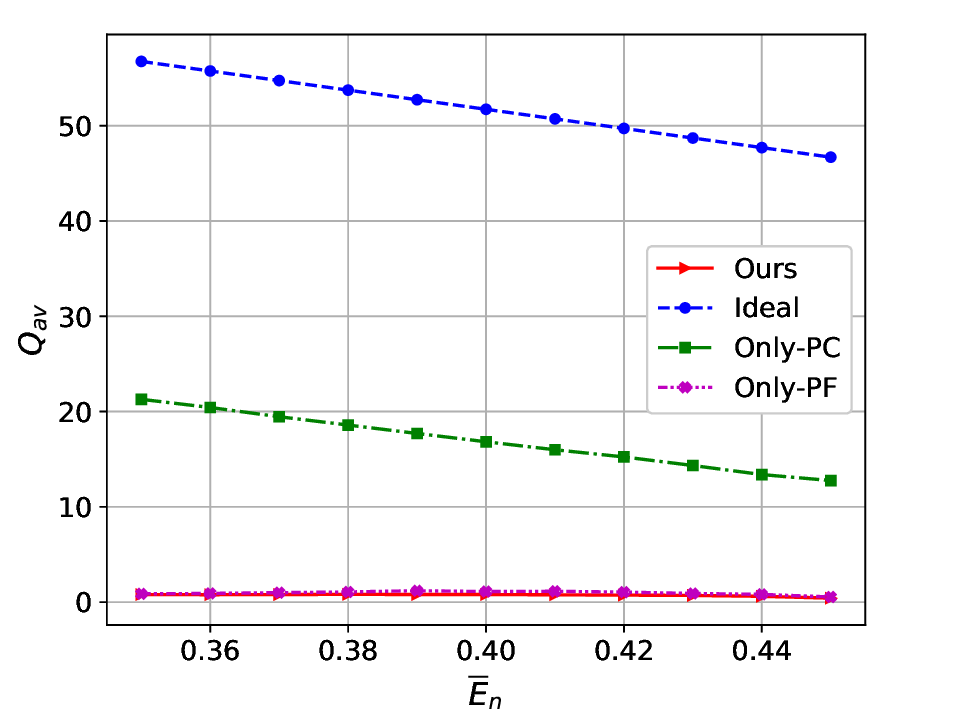}
}
% \vspace{-0.1in}
\caption{Impact of $\overline{E}_n$ on MNIST dataset.}
\label{fig: Impact of E_n MNIST}
\end{figure}

\begin{figure}[t]
\centering
% \subfigcapskip=-10pt
\subfigure[Average energy consumption vs. $\overline{E}_n$ on CIFAR-10 dataset.]{\label{fig:Average energy consumption vs E CIFAR-10}
    \includegraphics[clip, viewport= 5 0 420 330,width=0.45\linewidth]{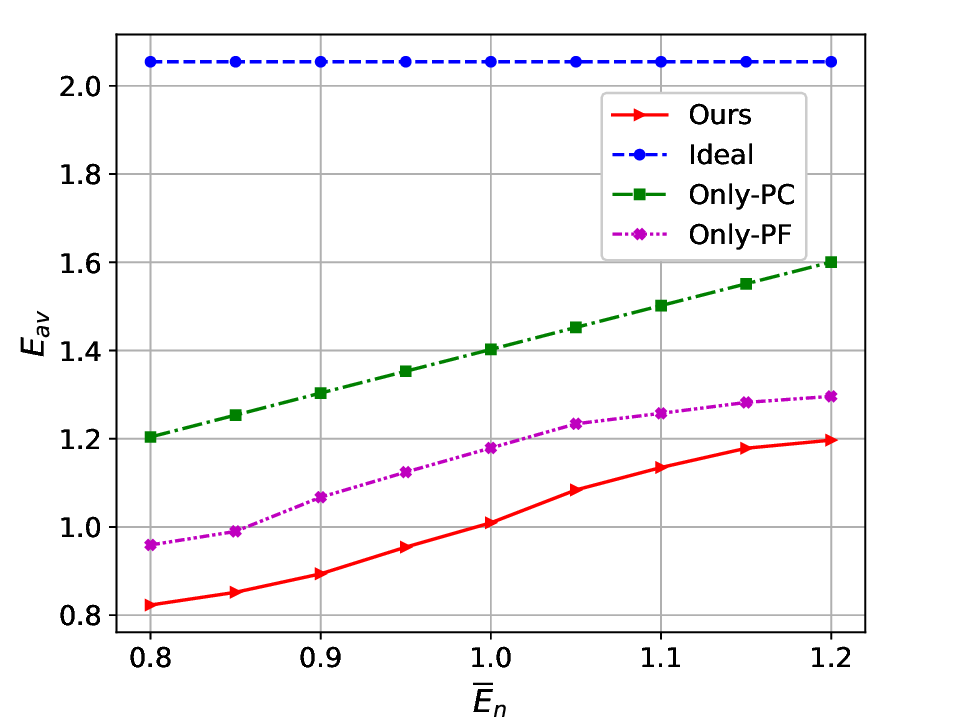}}
\ 
\subfigure[Average queue length vs. $\overline{E}_n$ on CIFAR-10 dataset.]{\label{fig: Average convergence error vs E CIFAR-10}
    \includegraphics[clip, viewport= 5 0 420 330,width=0.45 \linewidth]{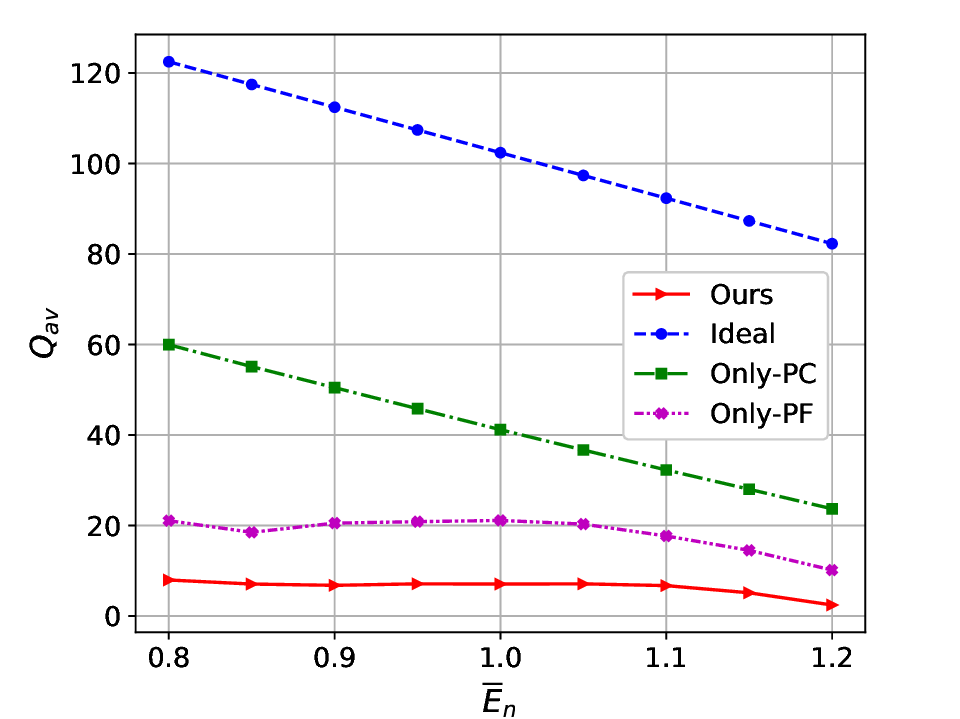}
}
% \vspace{-0.1in}
\caption{Impact of $\overline{E}_n$ on CIFAR-10 dataset.}
\label{fig: Impact of E_n CIFAR-10}
\end{figure}
Fig. \ref{fig: Impact of E_n MNIST} and Fig. \ref{fig: Impact of E_n CIFAR-10} show the impact of the energy consumption budget $\overline{E}_n$ on the average energy consumption and the virtual queue length of all schemes on both datasets. As expected, the average energy consumption increases with $\overline{E}_n$ for the proposed scheme, the Only-PC scheme and the Only-PF scheme. But in the Ideal FL scheme, the average energy consumption remains at a highest value. This is because the devices have to consume more energy to update and successfully upload the local gradients without freezing. For the other schemes, a higher energy consumption budget allows the devices to train the learning model with a smaller parameter freezing percentage or tolerate a higher transmit power for uploading local gradient parameters. Furthermore, the virtual queue length for all schemes decrease with $\overline{E}_n$, as more energy budget is available for training in each slot. Notably, the proposed scheme can achieve the smallest queue length compared to the other three benchmarks, which accounts for the stable energy consumption. 
\vspace{-0.2cm}
\subsection{Performance Comparison with the State-of-the-art Methods}
\vspace{-0.1cm}
We also adopt four state-of-the-art baselines for performance comparison. The baselines are summarized as follows. 
\begin{itemize} 
    \item Top-$K$ sparsification \cite{shi2019convergence}: Each device updates the entire local model and sends the top-$K$ most significant elements of the local gradient to the server, while the rest are accumulated locally. To be fair, we set $\gamma_n^k = 1-\frac{K_n^k}{D}$, where $D$ is the dimension of the local gradient, and set $\gamma_n^k$ to be the same as in the proposed scheme. 
    \item Model pruning \cite{9598845}: Each device evaluates the importance of a parameter by squaring the product of the corresponding global gradient and the parameter itself. Subsequently, the unimportant parameters are pruned. To ensure fairness, the pruning percentage of each device is set the same as the parameter freezing percentage of the proposed scheme.
    \item FedHQ \cite{9425020}: Each device adjusts the quantization policy according to its channel condition. To be fair, we set the size of transmission bits is to be the same as in the proposed scheme.  
    \item FjORD \cite{NEURIPS2021_6aed000a}: Each device extracts a sub-model from the global model using ordered dropout. The dropout percentage of each device is set the same as the parameter freezing percentage of the proposed scheme.
\end{itemize}
\begin{figure}[t]
\centering
\subfigure[CIFAR-10 dataset (non-IID).]{\label{fig:Test accuracy vs. total energy consumption CIFAR10 non IID sota}
    \includegraphics[clip, viewport= 3 3 432 335,width=0.45 \linewidth]{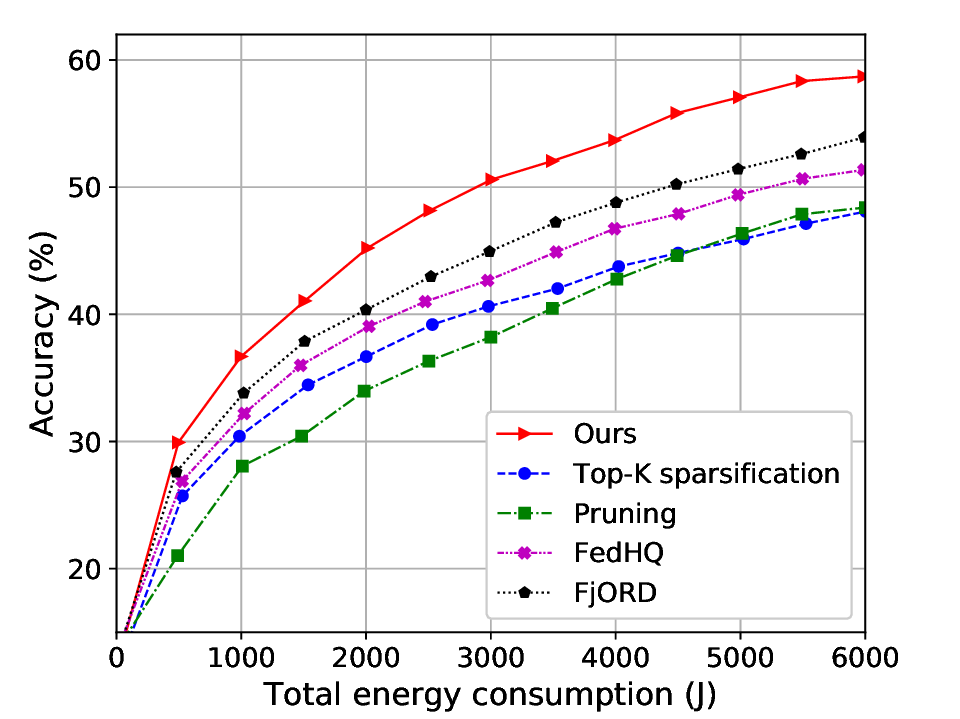}
}
\ 
\subfigure[CIFAR-10 dataset (IID).]{\label{fig:Test accuracy vs. total energy consumption CIFAR10 IID sota}
    \includegraphics[clip, viewport= 3 3 432 335,width=0.45 \linewidth]{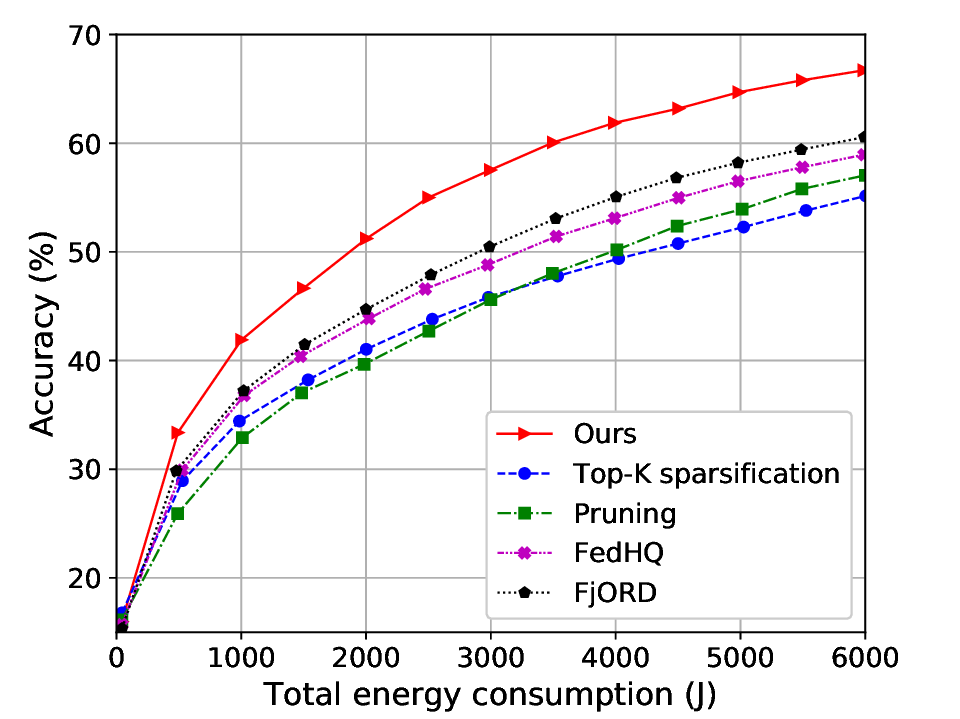}
}
\caption{Test accuracy vs. total energy consumption on CIFAR-10 dataset with IID and non-IID data. }
\label{fig:Test accuracy vs. total energy consumption sota}
\end{figure}\par
\begin{figure}[t]
\centering
\subfigure[CIFAR-10 dataset (non-IID).]{
    \includegraphics[clip, viewport= 3 3 432 335,width=0.45 \linewidth]{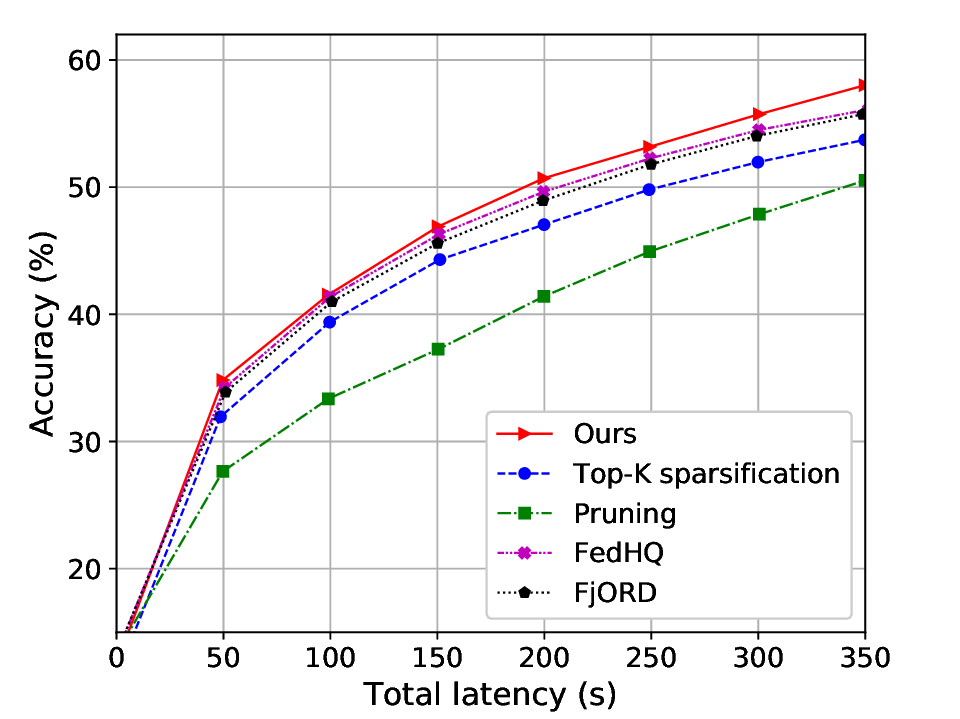}
}
\ 
\subfigure[CIFAR-10 dataset (IID).]{
    \includegraphics[clip, viewport= 3 3 432 335,width=0.45 \linewidth]{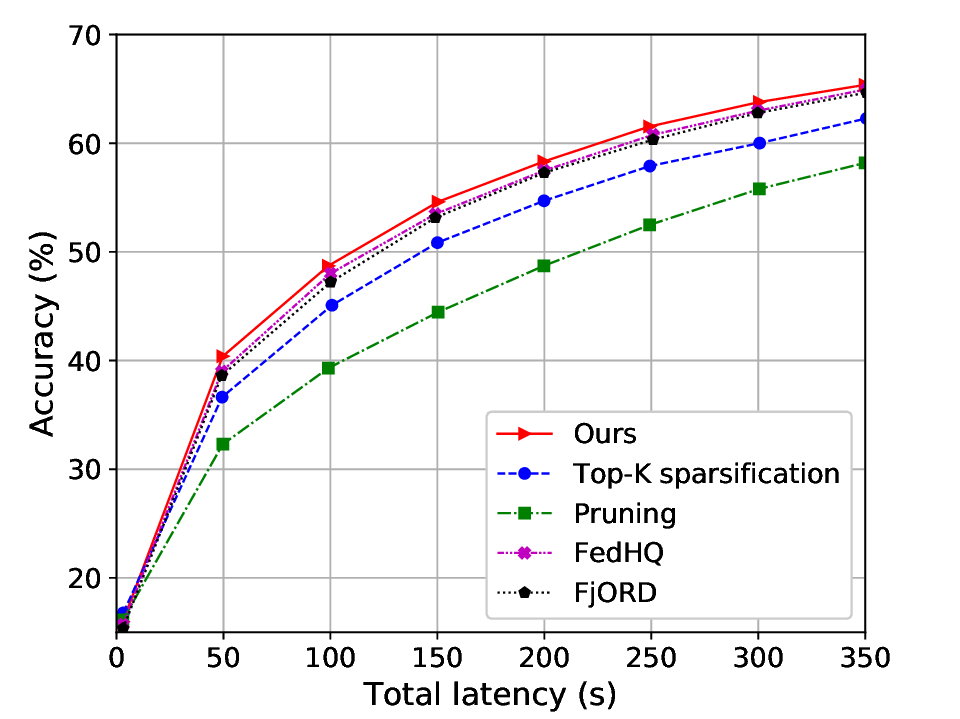}
}
% \vspace{-0.1in}
\caption{Test accuracy vs. total latency on CIFAR-10 dataset with IID and non-IID data. }
\label{fig:Test accuracy vs. total latency consumption sota}
\end{figure}\par
Fig. \ref{fig:Test accuracy vs. total energy consumption sota} shows the test accuracy of all schemes versus the total energy consumption of the devices on non-IID and IID data. As expected, the proposed scheme achieves higher accuracy than the other benchmarks at the same total energy consumption level. This is because the proposed scheme freezes the stable parameters from updating to save unnecessary computational energy consumption. In contrast, the top-$K$ sparsification scheme, the FedHQ scheme, and the Snowball scheme need to update the entire local model and thus leads to redundant computational energy consumption for updating stable parameters. The model pruning scheme and the FjORD scheme, which involve direct discarding of model parameters, damage the accuracy of the model and thus result in slow convergence. \par
Fig. \ref{fig:Test accuracy vs. total latency consumption sota} shows the test accuracy of all state-of-the-art schemes versus the total latency on non-IID and IID data. We observe that the proposed scheme outperforms both the top-K sparsification scheme and the FedHQ scheme. This is because the proposed scheme reduces the time required to update stable model parameters, allowing more devices to successfully transmit, thereby enhancing the overall model performance. In contrast, the top-K sparsification scheme and the FedHQ scheme require updating the entire set of model parameters, which increases computation latency across multiple devices, leading to a higher likelihood of transmission outages and ultimately degrading performance. Moreover, both the model pruning scheme and the FjORD scheme compromise model accuracy by directly discarding model parameters. 
\subsection{Justification of Assumption \ref{Assumption 4}}
\label{Justification of Assumption 3}
To further validate Assumption \ref{Assumption 4}, we conduct two experiments as follows. Specifically, MobileNetV2 and ResNet-20 are trained on the CIFAR-10 dataset using 30 devices. Subsequently, three devices are randomly selected, and the norm of the parameter gap induced by freezing is computed. 
\begin{figure}[H]
\centering
\subfigure[MobileNetV2.]{\label{fig_Norm_of_the_parameter_gap_MobileNetV2}
    \includegraphics[clip, viewport= 5 5 450 330,width=0.45 \linewidth]{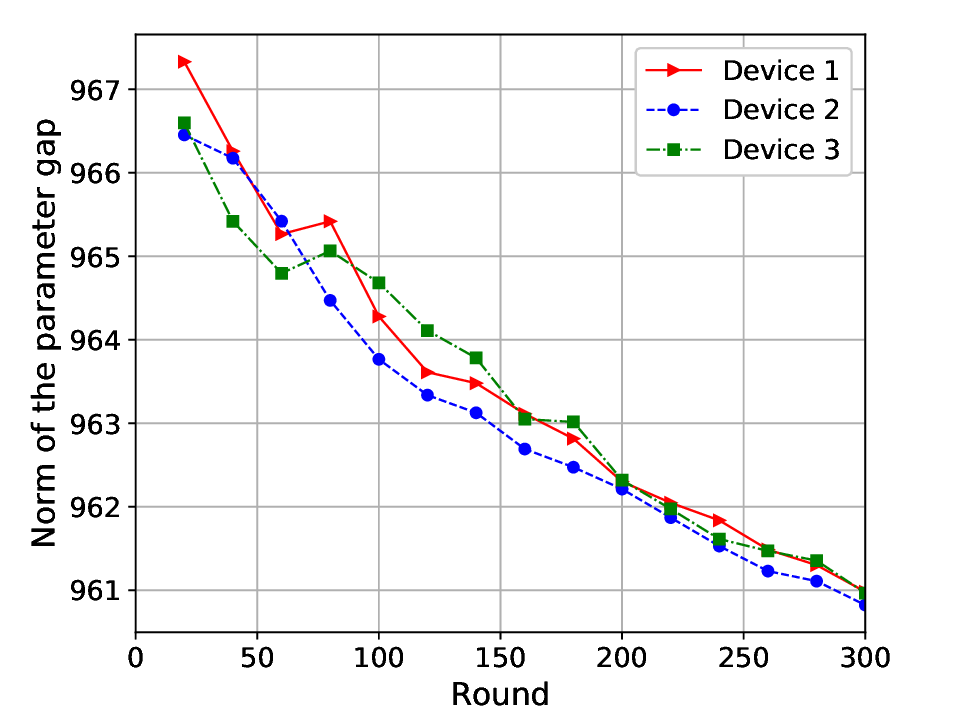}
}
\subfigure[ResNet-20.]{\label{fig_Norm_of_the_parameter_gap_ResNet20}
    \includegraphics[clip, viewport= 5 5 450 330,width=0.45 \linewidth]{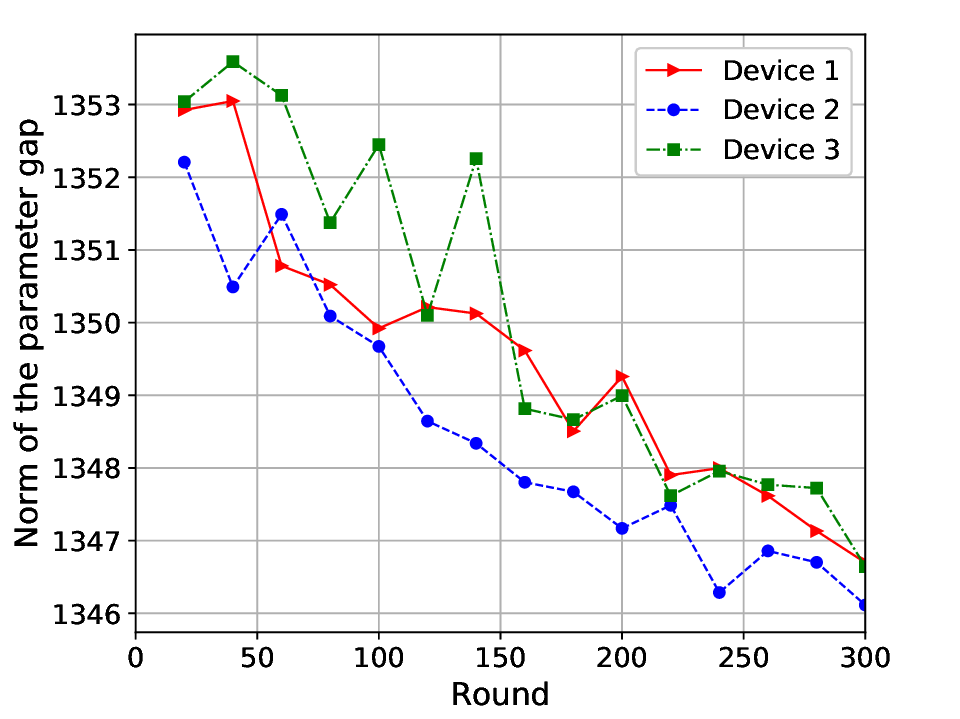}
}
\caption{The norm of parameter gap induced by
freezing on (a) MobileNetV2 and (b) ResNet-20. }
\label{fig_Norm_of_the_parameter_gap}
\end{figure}
\vspace{-0.5cm}\par
Fig. \ref{fig_Norm_of_the_parameter_gap} shows the norm of the parameter gap induced by freezing on three randomly selected devices during training with MobileNetV2 and ResNet-20 on the CIFAR-10 dataset. We observe that the norms remain bounded throughout the training process, thereby justifying Assumption \ref{Assumption 4}.

\section{Conclusion}
\label{sec:Conclusion}
In this paper, we focus on improving energy efficiency for deploying FL over wireless networks. We propose a two-timescale FL framework with joint parameter freezing and power control to further reduce energy consumption over wireless networks. Then we derive a convergence bound for the proposed FL scheme. Based on the convergence analysis, we formulate a problem with joint optimization of parameter freezing percentage and transmit power to minimize the convergence error of the learning model while ensuring the stability of energy consumption for each device. To solve the problem, a low-complexity online algorithm is developed. Comprehensive theoretical analysis and experimental results confirm the feasibility and superiority of the proposed scheme compared to the benchmark schemes.

\begin{appendix}

\subsection{Proof of Theorem 1}

\label{proof-of-theorem1}
To prove Theorem 1, the update function of global model can be rewritten as
\begin{align} \label{rewritten update}
    \bm{w}^{t+1} = \bm{w}^{t} - \eta (\nabla F(\bm{w}^t) - \bm{o}),
\end{align}
where $\bm{o} = \nabla F(\bm{w}^t) - g(\tilde{\bm{w}}^{t})$ is the global gradient bias introduced by parameter freezing, transmission outage, and data heterogeneity. 
Then in the $t$-th communication round, we have
\begin{align}
    &F(\bm{w}^{t+1}) - F(\bm{w}^{t}) \nonumber\\
    & \leq \langle \nabla{F}(\bm{w}^{t}), \bm{w}^{t+1}-\bm{w}^{t}\rangle +\frac{L}{2} \Vert\bm{w}^{t+1} - \bm{w}^{t}\Vert^2 \nonumber\\
    & = \langle \nabla{F}(\bm{w}^{t}), -\eta (\nabla F(\bm{w}^{t}) - \bm{o}) \rangle +\frac{L \eta^2}{2} \Vert \nabla F(\bm{w}^{t}) - \bm{o}\Vert^2.
\end{align}
Let $\eta = \frac{1}{L}$ and taking total expectation on both sides, we have 
\begin{align}\label{expectation expression}
    &\mathbb{E} \left [F(\bm{w}^{t+1}) - F(\bm{w}^{t}) \right] \nonumber\\
    &\leq  - \frac{1}{2L} \mathbb{E} \left [\Vert \nabla{F}(\bm{w}^{t}) \Vert^2\right] + \frac{1}{2L}\mathbb{E}\left[\Vert\bm{o}\Vert^2\right]. 
\end{align}
Then we can derive the $\mathbb{E}\left[\Vert\bm{o}\Vert^2\right]$ as follows 
\begin{align}\label{convergence error}
    &\mathbb{E}\left[\Vert\bm{o}\Vert^2\right] \nonumber\\
    &= \mathbb{E} \left[ \left \Vert \nabla{F}(\bm{w}^{t}) - \sum\limits_{n \in \mathcal{N}_1 \cup \mathcal{N}_2} \frac{B_{n}^t \nabla{F_{n}(\bm{w}^{t})}}{B^t} \right. \right. \nonumber\\ 
    &\left. \left. \quad + \sum\limits_{n \in \mathcal{N}_1 \cup \mathcal{N}_2} \frac{B_{n}^t \nabla{F_{n}(\bm{w}^{t})}}{B^t} -g(\tilde{\bm{w}}^{t}) \right \Vert^2\right] \nonumber\\
    & = \mathbb{E} \left[ \left \Vert \nabla{F}(\bm{w}^{t}) - \sum\limits_{n \in \mathcal{N}_1 \cup \mathcal{N}_2} \frac{B_{n}^t \nabla{F_{n}(\bm{w}^{t})}}{B^t} \right. \right. \nonumber\\ 
    &\left. \left. \quad + \sum\limits_{n \in \mathcal{N}_1} \frac{B_{n}^t \nabla{F_{n}(\bm{w}^{t})}}{B^t} \right. \right. \nonumber\\ 
    &\left. \left. \quad
    + \sum\limits_{n \in \mathcal{N}_2} \frac{B_{n}^t \nabla{F_{n}(\bm{w}^{t})}}{B^t}-g(\tilde{\bm{w}}^{t}) \right \Vert^2\right] \nonumber\\
    & \overset{(a_1)}{\leq} 3\mathbb{E} \left[ \left \Vert \nabla{F}(\bm{w}^{t}) - \sum\limits_{n \in \mathcal{N}_1 \cup \mathcal{N}_2} \frac{B_{n}^t \nabla{F_{n}(\bm{w}^{t})}}{B^t} \right\Vert^2\right] \nonumber\\ 
    & \quad + 3 \underbrace{\mathbb{E} \left[ \left \Vert\sum\limits_{n \in \mathcal{N}_1} \frac{B_{n}^t}{B^t} \nabla{F_{n}(\bm{w}^{t})}\right \Vert^2\right]}_{A_1}\nonumber\\ 
    & \quad + 3 \underbrace{\mathbb{E} \left[ \left \Vert \sum\limits_{n \in \mathcal{N}_2} \frac{B_{n}^t}{B^t} \nabla{F_{n}(\bm{w}^{t})} - g(\tilde{\bm{w}}^{t})\right \Vert^2\right]}_{A_2},
\end{align}
where $\mathcal{N}_1$ represents the set of devices that experience a transmission outage (i.e., $\mathbbm{1}_{n}^{t} = 0$), and $\mathcal{N}_2$ is the set of devices that successfully upload the local gradient parameters without freezing (i.e., $\mathbbm{1}_{n}^{t} = 1$). Moreover, inequality $(a_1)$ follows from the following inequality, which can be derived using the Cauchy-Schwarz inequality and the AM-GM inequality: $\|\mathbf{x}_1 + \mathbf{x}_2 + \mathbf{x}_3\|^2 \leq 3 (\|\mathbf{x}_1\|^2 + \|\mathbf{x}_2\|^2 + \|\mathbf{x}_3\|^2)$. 
Then according to the Assumption 4, we have
\begin{align}\label{A0}
    \mathbb{E} \left[ \left \Vert \nabla{F}(\bm{w}^{t}) - \sum\limits_{n=1}^N \frac{B_{n}^t \nabla{F_{n}(\bm{w}^{t})}}{B^t} \right\Vert^2\right] \overset{(a_2)}{\leq} \sum\limits_{n=1}^{N} \frac{B_{n}^t \mathcal{X}_n^2}{B^t},
\end{align}
where inequality $(a_2)$ is due to the Jensen's inequality. Moreover, according to the Assumption 2, we have
\begin{align}\label{A1}
    &A_1 = \mathbb{E} \left[ \left \Vert \sum_{n=1}^{N} \frac{(1-\mathbbm{1}_{n}^{t})B_{n}^t}{B^t} \nabla{F_n(\bm{w}^{t})} \right \Vert^2\right]\nonumber\\
    & \overset{(a_3)}{\leq} \left(\sum_{n=1}^{N}\frac{(1-\mathbbm{1}_{n}^{t})B_{n}^t}{B^t}\right)  \frac{\sum\limits_{n=1}^{N}(1-\mathbbm{1}_{n}^{t})B_{n}^t\mathbb{E} \left[ \left \Vert \nabla{F_n(\bm{w}^{t})}\right \Vert^2\right]}{\sum_{n=1}^{N}(1-\mathbbm{1}_{n}^{t})B_{n}^t}\nonumber\\
    & \overset{(b_3)}{\leq} \left(\sum_{n=1}^{N} \frac{(1-\mathbbm{1}_{n}^{t})B_{n}^t}{B^t}\right) \left(\xi_1+\xi_2\mathbb{E} \left[ \left \Vert \nabla{F(\bm{w}^{t})}\right \Vert^2\right]\right), 
\end{align}
where inequality $(a_3)$ is due to Jensen's inequality, and inequality $(b_3)$ follows Assumption 2. Moreover, we have
\begin{align}\label{A2}
    &A_2 = \mathbb{E} \left[ \left \Vert \sum_{n=1}^{N} \frac{\mathbbm{1}_{n}^{t} B_{n}^t}{B^t} \nabla{F_{n}(\bm{w}^{t})} - \frac{\sum_{n=1}^{N} \mathbbm {1}_{n}^{t} B_{n}^t  \nabla{F_{n}(\tilde{\bm{w}}_{n}^{t})}}{\sum_{n=1}^{N} \mathbbm {1}_{n}^{t} B_{n}^t }\right \Vert^2\right] \nonumber\\ 
    &\leq 2\mathbb{E}\left[ \left \Vert \frac{(B^t - \sum_{n=1}^{N} \mathbbm{1}_n^t B_{n}^t)\sum_{n=1}^{N} \mathbbm{1}_n^t B_{n}^t \nabla F_n(\tilde{\bm{w}}_n^t)}{B^t\sum_{n=1}^{N} \mathbbm{1}_n^t B_{n}^t}\right \Vert^2\right]  \nonumber\\& + 2\mathbb{E}\left[ \left \Vert \frac{(\sum\limits_{n=1}^{N} \mathbbm{1}_n^t B_{n}^t)\sum\limits_{n=1}^{N} \mathbbm{1}_n^t B_{n}^t \left(\nabla F_n(\bm{w}^{t})-\nabla F_n(\tilde{\bm{w}}_n^t)\right)}{B^t \sum_{n=1}^{N} \mathbbm{1}_n^t B_{n}^t}\right \Vert^2\right]\nonumber\\
    & \overset{(a_4)}{\leq} 2\left(\sum\limits_{n=1}^{N}  \frac{\mathbbm{1}_n^t B_{n}^t}{B^t}\right)\frac{\sum\limits_{n=1}^{N} \mathbbm{1}_n^t B_{n}^t\mathbb{E}\left[\left \Vert\nabla F_n(\bm{w}^{t})-\nabla F_n(\tilde{\bm{w}}_n^t)\right \Vert^2\right]}{\sum_{n=1}^{N} \mathbbm{1}_n^t B_{n}^t} \nonumber
    \\& \quad + 2\left(\sum_{n=1}^{N}\frac{(1 -\mathbbm{1}_n^t)B_{n}^t}{B^t}\right)\frac{\sum_{n=1}^{N} \mathbbm{1}_n^t B_{n}^t\mathbb{E}\left[\left \Vert\nabla F_n(\tilde{\bm{w}}_n^t)\right \Vert^2\right]}{\sum_{n=1}^{N} \mathbbm{1}_n^t B_{n}^t}\nonumber\\
    & \overset{(b_4)}{\leq} 2\left(\sum_{n=1}^{N}  \frac{\mathbbm{1}_n^t B_{n}^t}{B^t}\right)\frac{L^2 \sigma^2\sum_{n=1}^{N} \mathbbm{1}_n^t B_{n}^t \gamma_n^k}{\sum_{n=1}^{N} \mathbbm{1}_n^t B_{n}^t} \nonumber\\& + 2\left(\sum_{n=1}^{N}\frac{(1 -\mathbbm{1}_n^t)B_{n}^t}{B^t}\right)\left(\xi_1+\xi_2\mathbb{E} \left[ \left \Vert \nabla{F(\bm{w}^{t})}\right \Vert^2\right]\right), 
\end{align} 
where inequality $(a_4)$ follows Jensen's inequality, and inequality $(b_4)$ is due to the Assumption 2 and the fact that, 
\begin{align}
        &\mathbb{E}\left[\left \Vert\nabla F_n(\bm{w}^{t})-\nabla F_n(\tilde{\bm{w}}_n^t)\right \Vert^2\right]\overset{(a_5)}{\leq} L^2 \sigma^2 \gamma_n^k,
\end{align}
where inequality $(a_5)$ is due to the Assumption 1 and the Lemma 1.
Then plugging (\ref{A0}), (\ref{A1}), (\ref{A2}) and (\ref{convergence error}) back to (\ref{expectation expression}), we have
\begin{align}
    &\mathbb{E}[F(\bm{w}^{t+1})-F(\bm{w}^{t})] \nonumber\\
    &\leq - A^t \mathbb{E}[\Vert\nabla F(\bm{w}^{t})\Vert^2] + \frac{3}{2L} \sum_{n=1}^{N} \frac{B_{n}^t \mathcal{X}_n^2}{B^t}\nonumber\\
    &+ \frac{3L \sigma^2}{B^t} \sum\limits_{n=1}^{N} \mathbbm{1}_n^t B_{n}^t \gamma_n^k + \frac{9\xi_1}{2LB^t}\sum\limits_{n=1}^{N}(1-\mathbbm{1}_n^t)B_{n}^t, 
\end{align}
where $A^t = \frac{1}{2L} - \sum_{n=1}^{N} \frac{9\xi_2(1- \mathbbm{1}_n^t)B_{n}^t}{2LB^t}$. Let $\frac{1-9\xi_2}{2L} \leq A^t \leq \frac{1}{2L}$, we can obtain that 
\begin{align}\label{the expected convergence error}
    &\mathbb{E}[\Vert\nabla F(\bm{w}^{t})\Vert^2]\nonumber \\
    &\leq \frac{2L}{1-9\xi_2} \mathbb{E}\left[F(\bm{w}^t)-F(\bm{w}^{t+1})\right] + \underbrace{\frac{3}{(1-9\xi_2)B^t}\sum\limits_{n=1}^{N} B_n^t \mathcal{X}_n^2}_{\textit{error caused by data heterogeneity}}\nonumber \\
    &+ \underbrace{\frac{6L^2 \sigma^2}{(1-9\xi_2)B^t}\sum\limits_{n=1}^{N} \mathbbm{1}_n^t B_{n}^t \gamma_n^k}_{X_1^t \textit{ caused by parameter freezing}} +\underbrace{\frac{9\xi_1}{(1-9\xi_2)B^t}\sum\limits_{n=1}^{N} (1-\mathbbm{1}_n^t) B_{n}^t}_{X_2^t \textit{ caused by transmission outage}}, 
\end{align}
where $B^t = \sum_{n=1}^{N} B_{n}^t$ is the total data size of all devices.
This completes the proof.

\subsection{Proof of Corollary 2}\label{proof-of-corollary-2}
To investigate the impact of outage probability on convergence performance, we first recall that the channel power gains in each slot are modeled as $h_n^t = g_n^t H_n^t$, where $H_n^t$ is the large-scale fading, and the small-scale fading $g_n^t$ follows normalized exponential distribution. Then we have $h_n^t \sim \exp\left(\frac{1}{H_n^t}\right), \forall n \in \mathcal{N}$. Then according to Definition 1, a transmission outage occurs when the sum of communication latency $\tau_{n}^{{\rm{com}}, t}$ and computation latency $\tau_{n}^{{\rm{cmp}}, t}$ exceeds a  given per-round latency $\tau_0$; that is, the outage probability is given by
\begin{align}
    q_n^t &\triangleq {\rm{Pr}}\{ \tau_{n}^{{\rm{cmp}}, t}+\tau_{n}^{{\rm{com}}, t} > \tau_0\} = 1 - e^{-\frac{h_{n, \min}^t}{H_n^t}}, 
\end{align}
where $h_{n, \min}^t = \frac{N_0}{p_n^t}\left( 2^{\frac{(1-\gamma_n^k)S}{W(\tau_0 - \tau_n^{{\rm{cmp}}, t})}} - 1\right)$, and $\tau_{n}^{{\rm cmp},t} = \frac{(1-\gamma_n^k)c_n B_{n}^t}{f_n}$. \par 
Taking the expectation with respect to channel randomness, we have $\mathbb{E}_{h_n^t} [\mathbbm{1}_n^t] = 1 - q_n^t$. Plugging it back to (\ref{the expected convergence error}), we complete the proof. 

\subsection{Proof of Lemma 2}
\label{proof of lemma drift-plus-penalty upper bound 1}
According to the queue dynamics, we have
\begin{align}\label{queue update}
    Q_n^{t+1} = \left[Q_n^t+E_n^t-\overline{E}_n\right]^{+}, \forall n \in \mathcal{N}, t = 0, 1, \cdots
\end{align}\par
Then we can obtain that
\begin{align}\label{queue dynamics}
    (Q_n^{t+1})^2 - (Q_n^t)^2 &\overset{(a_5)}{\leq} \left[Q_n^t+E_n^t-\overline{E}_n\right]^2 -(Q_n^t)^2\nonumber\\
    &\leq (E_n^t)^2+\overline{E}_n^2+2Q_n^t(E_n^t-\overline{E}_n), \forall n \in \mathcal{N},
\end{align}
where $Q_n^{t+1} = \max\left\{Q_n^t+E_n^t-\overline{E}_n,0\right\}$, $(a_5)$ is due to $\max\left\{x,0 \right\}^2 \leq x^2$. Summing the above (\ref{queue dynamics}) over $t \in \mathcal{T}_k$, the conditional Lyapunov drift $\Delta_{n,T}(Q_n^{kT})$ is upper bounded by
\begin{align}\label{drift 2}
    & \Delta_{n,T}(Q_n^{kT}) \nonumber\\
    & = \mathbb{E}\{ \frac{1}{2} (Q_n^{(k+1)T})^2 - \frac{1}{2}(Q_n^{kT})^2 \Big| Q_n^{kT}\}\nonumber\\
    & \leq \mathbb{E} \{\frac{1}{2} \sum_{t\in \mathcal{T}_k} (E_n^t)^2 + \frac{1}{2}\overline{E}_n^2 T  +\sum_{t\in \mathcal{T}_k} Q_n^t(E_n^t-\overline{E}_n) \Big| Q_n^{kT} \}\nonumber\\
    & \leq \beta_1 T +\mathbb{E}\{ \sum_{t\in \mathcal{T}_k} Q_n^t(E_n^t-\overline{E}_n) \Big| Q_n^{kT}\}, 
\end{align}
where $\beta_1 = \frac{1}{2}(E_{n, \max}^2+\overline{E}_n^2)$, $E_{n, \max} = \max\{E_n^t\}, t \in \mathcal{T}_k$. Adding the term $V \mathbb{E}\left\{ \sum_{t \in \mathcal{T}_k} X_n^t \mid Q_n^{kT} \right\}$ into both sides of (\ref{drift 2}), we prove the Lemma 2. 

\subsection{Proof of Lemma 3}
\label{proof of lemma drift-plus-penalty upper bound 2}
According to (\ref{queue update}) and $E_{n, \max} \geq E_n(t), \forall n \in \mathcal{N}$, we have
\begin{align}
    Q_n^{kT}-(t-kT)\overline{E}_n \leq Q_n^t \leq Q_n^{kT}+(t-kT)(E_{n, \max} - \overline{E}_n). 
\end{align}\par
Then the term $\sum_{t \in \mathcal{T}_k} Q_n^t(E_n^t-\overline{E}_n) $ can be bounded as
\begin{align}\label{energy queue term}
    &\sum_{t \in \mathcal{T}_k} Q_n^t(E_n^t-\overline{E}_n)\nonumber\\
    &\leq \sum_{t \in \mathcal{T}_k} Q_n^{kT}(E_n^t-\overline{E}_n) \nonumber\\ 
    & \quad + \sum_{t \in \mathcal{T}_k} (t-kT)\left[(E_{n, \max}-\overline{E}_n)E_n^t + \overline{E}_n^2\right].
\end{align}\par
Taking the conditional expectation on (\ref{energy queue term}), we have
\begin{align}
    &\mathbb{E}\{ \sum_{t \in \mathcal{T}_k} Q_n^t(E_n^t-\overline{E}_n) \Big| Q_n^{kT}\} \nonumber\\
    & \leq  \mathbb{E}\{ \sum_{t \in \mathcal{T}_k} Q_n^{kT}(E_n^t-\overline{E}_n) \Big| Q_n^{kT}\} \nonumber\\
    & \quad + \mathbb{E}\{ \sum_{t \in \mathcal{T}_k} (t-kT)[(E_{n, \max}-\overline{E}_n)E_n^t + \overline{E}_n^2] \}\nonumber\\
    & \leq \mathbb{E}\{ \sum_{t \in \mathcal{T}_k} Q_n^{kT}(E_n^t-\overline{E}_n) \Big| Q_n^{kT}\} + \beta_3, 
\end{align}
where $\beta_3 = \frac{T(T-1)[(E_{n, \max}-\overline{E}_n)E_{n, \max} + \overline{E}_n^2]}{2}$. Let $ \beta_2 = \beta_1 + \beta_3 / T$, we can prove the Lemma 3.

\subsection{Proof of Proposition 2}
\label{proof of Proposition 2}

Accordingly, we have $\tilde{Z}''(\gamma \mid \gamma^l)\geq 0$, $\tilde{Z}(\gamma^l \mid \gamma^l) = {Z}(\gamma^l)$, and $\tilde{Z}'(\gamma^l \mid \gamma^l) = {Z}'(\gamma^l)$. Thus the function $\tilde{Z}''(\gamma \mid \gamma^l)$ is convex and locally equal at the feasible point $\gamma^l$. Moreover, the inequality $\tilde{Z}(\gamma \mid \gamma^l) \geq {Z}(\gamma)$ holds when $M \geq Z''(\gamma), \forall \gamma \in [0, 1]$. To obtain the upper bound of $Z''(\gamma)$, i.e., $M$, we have the following discussion. For notational brevity, the superscript $k$ and the subscript $n$ are omitted. Then we have\par
Case 1: When $p_{\max} < \overline{P}$, we can obtain that 
\begin{align}
    h_{\min}(\gamma) = \frac{C_1(\tau_0 - \theta(1-\gamma))}{I(1-\gamma)}\left( 2^{\frac{S(1-\gamma)}{W(\tau_{0} - \theta(1-\gamma))}}-1\right), 
\end{align}
where $C_1 = N_0Q, \Gamma \leq \gamma \leq 1, \theta = \frac{c_nB_n}{f_n}$, $\Gamma =1- \frac{\overline{P}Q\tau_0}{I+\overline{P}Q\theta} < 1$, and $I = VB\lambda-Qe^{\rm{cmp}}$. Combined with $\gamma \in [0, 1]$, we can obtain the range of $\gamma$ is $[\max\{0, \Gamma\}, 1]$. \par 
Then we can derive that $h_{\min}(\gamma)>0, h'_{\min}(\gamma)<0, h''_{\min}(\gamma)>0,$ and $h'''_{\min}(\gamma)<0, \forall \gamma \in [0, 1]$. 
And we have 
\begin{align}
    Z(\gamma) = I(\gamma -1) \Psi(\frac{h_{\min}(\gamma)}{H}), 
\end{align}
where $\phi(x) = \int_{x}^{\infty} \frac{e^{-t}}{t}dt, \Psi(x) = e^{-x}-x\phi(x)$, with $x>0$. Moreover $Z^{''}(\gamma)$ is bounded as 
\begin{align}
    &Z''(\gamma) 
    {\leq}I\underbrace{\frac{e^{-\frac{h_{\min}(\gamma)}{H}}}{h_{\min}(\gamma)}}_{G(\gamma)}[\underbrace{-2 h'_{\min}(\gamma) +(1-\gamma)h''_{\min}(\gamma)}_{Y(\gamma)}]. 
\end{align}
We can derive that $G(\gamma), G'(\gamma), Y(\gamma)>0, Y'(\gamma)<0, \forall \gamma \in [0, 1]$. Then we can obtain that
\begin{align}
M &= 
\begin{cases}
    I G(1)Y(0), &\text{if } \Gamma \leq 0; \\
    I G(1)Y(\Gamma), &\text{otherwise}.
\end{cases}
\end{align} \par 
Case 2: When $p_{\max} \geq \overline{P}$, we have
\begin{align}
    h_{\min}(\gamma) = \frac{N_0}{\overline{P}}\left( 2^{\frac{S(1-\gamma)}{W(\tau_0 - \theta(1-\gamma))}}-1\right), 
\end{align}
where $0 \leq \gamma \leq \Gamma$. We can derive that $h_{\min}(\gamma)>0, h'_{\min}(\gamma)<0, h''_{\min}(\gamma)>0,$ and $h'''_{\min}(\gamma)<0, \forall \gamma \in [0, \Gamma]$. 
Then $Z(\gamma)$ can be expressed as 
\begin{align}
    &Z(\gamma) = I(\gamma-1)e^{-\frac{h_{\min}(\gamma)}{H}} \nonumber\\&+ \overline{P}_nQ(\tau_0 -\theta(1-\gamma))\frac{h_{\min}(\gamma)}{H}\phi\left(\frac{h_{\min}(\gamma)}{H}\right). 
\end{align}
Similar to the Case 1, we have
\begin{align}
    M= &e^{-\frac{h_{\min}(\Gamma)}{H}}\left[\left( 2I+\overline{P}Q \theta\right)\left( -\frac{h'_{\min}(0)}{H}\right) + I\frac{h''_{\min}(0)}{H}\right].  
\end{align}\par
We complete the proof.\par
\end{appendix}

\bibliographystyle{IEEEtran}
\bibliography{IEEEabrv,link}

\begin{thebibliography}{10}
\providecommand{\url}[1]{#1}
\csname url@samestyle\endcsname
\providecommand{\newblock}{\relax}
\providecommand{\bibinfo}[2]{#2}
\providecommand{\BIBentrySTDinterwordspacing}{\spaceskip=0pt\relax}
\providecommand{\BIBentryALTinterwordstretchfactor}{4}
\providecommand{\BIBentryALTinterwordspacing}{\spaceskip=\fontdimen2\font plus
\BIBentryALTinterwordstretchfactor\fontdimen3\font minus \fontdimen4\font\relax}
\providecommand{\BIBforeignlanguage}[2]{{%
\expandafter\ifx\csname l@#1\endcsname\relax
\typeout{** WARNING: IEEEtran.bst: No hyphenation pattern has been}%
\typeout{** loaded for the language `#1'. Using the pattern for}%
\typeout{** the default language instead.}%
\else
\language=\csname l@#1\endcsname
\fi
#2}}
\providecommand{\BIBdecl}{\relax}
\BIBdecl

\bibitem{10310213}
X.~Liu, Y.~Deng, A.~Nallanathan, and M.~Bennis, ``Federated learning and meta learning: Approaches, applications, and directions,'' \emph{IEEE Commun. Surv. Tutor.}, vol.~26, no.~1, pp. 571--618, Fourth Quarter 2024.

\bibitem{8755300}
M.~Chen, U.~Challita, W.~Saad, C.~Yin, and M.~Debbah, ``Artificial neural networks-based machine learning for wireless networks: A tutorial,'' \emph{IEEE Commun. Surv. Tutor.}, vol.~21, no.~4, pp. 3039--3071, Fourth Quarter 2019.

\bibitem{8743390}
Y.~Sun, M.~Peng, Y.~Zhou, Y.~Huang, and S.~Mao, ``Application of machine learning in wireless networks: Key techniques and open issues,'' \emph{IEEE Commun. Surv. Tutor.}, vol.~21, no.~4, pp. 3072--3108, Fourth Quarter 2019.

\bibitem{pmlr-v54-mcmahan17a}
B.~McMahan, E.~Moore, D.~Ramage, S.~Hampson, and B.~A.~y. Arcas, ``Communication-efficient learning of deep networks from decentralized data,'' in \emph{Proc. Int. Conf. Artif. Intell. Stat. (AISTATS)}, vol.~54, Apr. 2017, pp. 1273--1282.

\bibitem{9562559}
M.~Chen, D.~Gündüz, K.~Huang, W.~Saad, M.~Bennis, A.~V. Feljan, and H.~V. Poor, ``Distributed learning in wireless networks: Recent progress and future challenges,'' \emph{IEEE J. Sel. Areas Commun.}, vol.~39, no.~12, pp. 3579--3605, Dec. 2021.

\bibitem{10304624}
C.~Xu, J.~Li, Y.~Liu, Y.~Ling, and M.~Wen, ``Accelerating split federated learning over wireless communication networks,'' \emph{IEEE Trans. Wireless Commun.}, vol.~23, no.~6, pp. 5587--5599, Jun. 2024.

\bibitem{9534784}
H.~Wang, Z.~Qu, Q.~Zhou, H.~Zhang, B.~Luo, W.~Xu, S.~Guo, and R.~Li, ``A comprehensive survey on training acceleration for large machine learning models in {IoT},'' \emph{IEEE Internet Things J.}, vol.~9, no.~2, pp. 939--963, Jan. 2022.

\bibitem{10185584}
Y.~Mao, Z.~Zhao, M.~Yang, L.~Liang, Y.~Liu, W.~Ding, T.~Lan, and X.-P. Zhang, ``Safari: Sparsity-enabled federated learning with limited and unreliable communications,'' \emph{IEEE Trans. Mob. Comput.}, vol.~23, no.~5, pp. 4819--4831, May 2024.

\bibitem{10233012}
Y.~Xu, Z.~Jiang, H.~Xu, Z.~Wang, C.~Qian, and C.~Qiao, ``Federated learning with client selection and gradient compression in heterogeneous edge systems,'' \emph{IEEE Trans. Mob. Comput.}, vol.~23, no.~5, pp. 5446--5461, May 2024.

\bibitem{8889996}
F.~Sattler, S.~Wiedemann, K.-R. Müller, and W.~Samek, ``Robust and communication-efficient federated learning from non-{IID} data,'' \emph{IEEE Trans. Neural Netw. Learn. Syst.}, vol.~31, no.~9, pp. 3400--3413, Sep. 2020.

\bibitem{NIPS2017_6c340f25}
D.~Alistarh, D.~Grubic, J.~Li, R.~Tomioka, and M.~Vojnovic, ``{QSGD}: Communication-efficient {SGD} via gradient quantization and encoding,'' in \emph{Proc. Adv. Neural Inf. Process. Syst. (NeurIPS)}, vol.~30, Dec. 2017, pp. 1709--1720.

\bibitem{9916128}
R.~Chen, L.~Li, K.~Xue, C.~Zhang, M.~Pan, and Y.~Fang, ``Energy efficient federated learning over heterogeneous mobile devices via joint design of weight quantization and wireless transmission,'' \emph{IEEE Trans. Mob. Comput.}, vol.~22, no.~12, pp. 7451--7465, Dec. 2023.

\bibitem{9272666}
G.~Zhu, Y.~Du, D.~Gündüz, and K.~Huang, ``One-bit over-the-air aggregation for communication-efficient federated edge learning: Design and convergence analysis,'' \emph{IEEE Trans. Wireless Commun.}, vol.~20, no.~3, pp. 2120--2135, Mar. 2021.

\bibitem{10319317}
Z.~Zhao, Y.~Mao, Z.~Shi, Y.~Liu, T.~Lan, W.~Ding, and X.-P. Zhang, ``{AQUILA}: Communication efficient federated learning with adaptive quantization in device selection strategy,'' \emph{IEEE Trans. Mob. Comput.}, vol.~23, no.~6, pp. 7363--7376, Jun. 2024.

\bibitem{10285619}
Q.~Pan, H.~Cao, Y.~Zhu, J.~Liu, and B.~Li, ``Contextual client selection for efficient federated learning over edge devices,'' \emph{IEEE Trans. Mob. Comput.}, vol.~23, no.~6, pp. 6538--6548, Jun. 2024.

\bibitem{9210812}
M.~Chen, Z.~Yang, W.~Saad, C.~Yin, H.~V. Poor, and S.~Cui, ``A joint learning and communications framework for federated learning over wireless networks,'' \emph{IEEE Trans. Wireless Commun.}, vol.~20, no.~1, pp. 269--283, Jan. 2021.

\bibitem{10038486}
J.~Yang, Y.~Liu, and R.~Kassab, ``Client selection for federated bayesian learning,'' \emph{IEEE J. Sel. Areas Commun.}, vol.~41, no.~4, pp. 915--928, Apr. 2023.

\bibitem{10086671}
Z.~Xu, D.~Li, W.~Liang, W.~Xu, Q.~Xia, P.~Zhou, O.~F. Rana, and H.~Li, ``Energy or accuracy? {Near}-optimal user selection and aggregator placement for federated learning in {MEC},'' \emph{IEEE Trans. Mob. Comput.}, vol.~23, no.~3, pp. 2470--2485, Mar. 2024.

\bibitem{10036106}
C.~Chen, H.~Xu, W.~Wang, B.~Li, B.~Li, L.~Chen, and G.~Zhang, ``Synchronize only the immature parameters: Communication-efficient federated learning by freezing parameters adaptively,'' \emph{IEEE Trans. Parallel Distrib. Syst.}, vol.~35, no.~7, pp. 1155--1173, Jul. 2024.

\bibitem{9829183}
Z.~Liang, Y.~Liu, T.-M. Lok, and K.~Huang, ``A two-timescale approach to mobility management for multicell mobile edge computing,'' \emph{IEEE Trans. Wireless Commun.}, vol.~21, no.~12, pp. 10\,981--10\,995, Dec. 2022.

\bibitem{10313248}
X.~Liu, S.~Wang, Y.~Deng, and A.~Nallanathan, ``Adaptive federated pruning in hierarchical wireless networks,'' \emph{IEEE Trans. Wireless Commun.}, vol.~23, no.~6, pp. 5985--5999, Jun. 2024.

\bibitem{10678894}
X.~Liu, T.~Ratnarajah, M.~Sellathurai, and Y.~C. Eldar, ``Adaptive model pruning and personalization for federated learning over wireless networks,'' \emph{IEEE Trans. Signal Process.}, vol.~72, pp. 4395--4411, Sep. 2024.

\bibitem{xie2024federated}
S.~Xie, D.~Wen, X.~Liu, C.~You, T.~Ratnarajah, and K.~Huang, ``Federated dropout: Convergence analysis and resource allocation,'' \emph{arXiv preprint arXiv:2501.00379}, 2024.

\bibitem{NEURIPS2021_6aed000a}
S.~Horv\'{a}th, S.~Laskaridis, M.~Almeida, I.~Leontiadis, S.~Venieris, and N.~Lane, ``{FjORD}: Fair and accurate federated learning under heterogeneous targets with ordered dropout,'' in \emph{Proc. Adv. Neural Inf. Process. Syst. (NeurIPS)}, vol.~34, Dec. 2021, pp. 12\,876--12\,889.

\bibitem{NEURIPS2022_bf5311df}
S.~Alam, L.~Liu, M.~Yan, and M.~Zhang, ``{FedRolex}: Model-heterogeneous federated learning with rolling sub-model extraction,'' in \emph{Proc. Adv. Neural Inf. Process. Syst. (NeurIPS)}, vol.~35, Dec. 2022, pp. 29\,677--29\,690.

\bibitem{NEURIPS2023_52635645}
H.~Zhou, T.~Lan, G.~P. Venkataramani, and W.~Ding, ``Every parameter matters: Ensuring the convergence of federated learning with dynamic heterogeneous models reduction,'' in \emph{Proc. Adv. Neural Inf. Process. Syst. (NeurIPS)}, vol.~36, Dec. 2023, pp. 25\,991--26\,002.

\bibitem{9673130}
R.~Jin, X.~He, and H.~Dai, ``Communication efficient federated learning with energy awareness over wireless networks,'' \emph{IEEE Trans. Wireless Commun.}, vol.~21, no.~7, pp. 5204--5219, Jul. 2022.

\bibitem{10443546}
B.~Luo, W.~Xiao, S.~Wang, J.~Huang, and L.~Tassiulas, ``Adaptive heterogeneous client sampling for federated learning over wireless networks,'' \emph{IEEE Trans. Mob. Comput.}, vol.~23, no.~10, pp. 9663--9677, Oct. 2024.

\bibitem{gu2021fast}
X.~Gu, K.~Huang, J.~Zhang, and L.~Huang, ``Fast federated learning in the presence of arbitrary device unavailability,'' in \emph{Proc. Adv. Neural Inf. Process. Syst. (NeurIPS)}, vol.~34, Dec. 2021, pp. 12\,052--12\,064.

\bibitem{10535531}
L.~Yu and T.~Ji, ``Efficient federated learning with channel status awareness and devices’ personal touch,'' \emph{IEEE Trans. Mob. Comput.}, vol.~23, no.~12, pp. 11\,794--11\,806, Dec. 2024.

\bibitem{10050151}
Z.~Jiang, Y.~Xu, H.~Xu, Z.~Wang, J.~Liu, Q.~Chen, and C.~Qiao, ``Computation and communication efficient federated learning with adaptive model pruning,'' \emph{IEEE Trans. Mob. Comput.}, vol.~23, no.~3, pp. 2003--2021, Mar. 2024.

\bibitem{9611373}
Y.~Wang, Y.~Xu, Q.~Shi, and T.-H. Chang, ``Quantized federated learning under transmission delay and outage constraints,'' \emph{IEEE J. Sel. Areas Commun.}, vol.~40, no.~1, pp. 323--341, Jan. 2022.

\bibitem{9598845}
S.~Liu, G.~Yu, R.~Yin, J.~Yuan, L.~Shen, and C.~Liu, ``Joint model pruning and device selection for communication-efficient federated edge learning,'' \emph{IEEE Trans. Commun.}, vol.~70, no.~1, pp. 231--244, Jan. 2022.

\bibitem{neely2010stochastic}
M.~J. Neely, ``Stochastic network optimization with application to communication and queueing systems,'' \emph{Synth. Lect. Commun. Netw.}, vol.~3, no.~1, pp. 1--211, 2010.

\bibitem{9151375}
S.~Wang, Y.-C. Wu, M.~Xia, R.~Wang, and H.~V. Poor, ``Machine intelligence at the edge with learning centric power allocation,'' \emph{IEEE Trans. Wireless Commun.}, vol.~19, no.~11, pp. 7293--7308, Nov. 2020.

\bibitem{boyd2004convex}
S.~Boyd and L.~Vandenberghe, \emph{Convex optimization}.\hskip 1em plus 0.5em minus 0.4em\relax Cambridge university press, 2004.

\bibitem{10346989}
J.~Yang, Y.~Liu, F.~Chen, W.~Chen, and C.~Li, ``Asynchronous wireless federated learning with probabilistic client selection,'' \emph{IEEE Trans. Wireless Commun.}, vol.~23, no.~7, pp. 7144--7158, Jul. 2024.

\bibitem{zeng2021energy}
Q.~Zeng, Y.~Du, K.~Huang, and K.~K. Leung, ``Energy-efficient resource management for federated edge learning with {CPU-GPU} heterogeneous computing,'' \emph{IEEE Trans. Wireless Commun.}, vol.~20, no.~12, pp. 7947--7962, Dec. 2021.

\bibitem{9593178}
W.~Shi, Y.~Sun, S.~Zhou, and Z.~Niu, ``Device scheduling and resource allocation for federated learning under delay and energy constraints,'' in \emph{Proc. 2021 IEEE 22nd Int. Workshop on Signal Process. Adv. in Wireless Commun. (SPAWC)}, Sep. 2021, pp. 596--600.

\bibitem{10089235}
Z.~Chen, W.~Yi, Y.~Liu, and A.~Nallanathan, ``Knowledge-aided federated learning for energy-limited wireless networks,'' \emph{IEEE Trans. Commun.}, vol.~71, no.~6, pp. 3368--3386, Jun. 2023.

\bibitem{yurochkin2019bayesian}
M.~Yurochkin, M.~Agarwal, S.~Ghosh, K.~Greenewald, N.~Hoang, and Y.~Khazaeni, ``Bayesian nonparametric federated learning of neural networks,'' in \emph{Proc. Int. Conf. Mach. Learn. (ICML)}.\hskip 1em plus 0.5em minus 0.4em\relax PMLR, 2019, pp. 7252--7261.

\bibitem{9237168}
J.~Xu and H.~Wang, ``Client selection and bandwidth allocation in wireless federated learning networks: A long-term perspective,'' \emph{IEEE Trans. Wireless Commun.}, vol.~20, no.~2, pp. 1188--1200, Feb. 2021.

\bibitem{sandler2018mobilenetv2}
M.~Sandler, A.~Howard, M.~Zhu, A.~Zhmoginov, and L.-C. Chen, ``{MobileNetv2}: Inverted residuals and linear bottlenecks,'' in \emph{Proc. IEEE Conf. Comput. Vis. Pattern Recogn.}, Jun. 2018, pp. 4510--4520.

\bibitem{9609994}
K.~Wei, J.~Li, C.~Ma, M.~Ding, C.~Chen, S.~Jin, Z.~Han, and H.~V. Poor, ``Low-latency federated learning over wireless channels with differential privacy,'' \emph{IEEE J. Sel. Areas Commun.}, vol.~40, no.~1, pp. 290--307, Jan. 2022.

\bibitem{shi2019convergence}
S.~Shi, K.~Zhao, Q.~Wang, Z.~Tang, and X.~Chu, ``A convergence analysis of distributed sgd with communication-efficient gradient sparsification.'' in \emph{Proc. 28th Int. Joint Conf. Artif. Intell. (IJCAI)}, Aug. 2019, pp. 3411--3417.

\bibitem{9425020}
S.~Chen, C.~Shen, L.~Zhang, and Y.~Tang, ``Dynamic aggregation for heterogeneous quantization in federated learning,'' \emph{IEEE Trans. Wireless Commun.}, vol.~20, no.~10, pp. 6804--6819, Oct. 2021.

\end{thebibliography}

\end{document}